\documentclass[10pt,twocolumn,letterpaper]{article}

\usepackage[pagenumbers,wacv]{wacv}      

\definecolor{wacvblue}{rgb}{0.21,0.49,0.74}
\usepackage[pagebackref,breaklinks,colorlinks,allcolors=wacvblue]{hyperref}
\usepackage[acronym]{glossaries}
\makenoidxglossaries%
\newacronym[plural={VLMs}]{vlm}{VLM}{Visual-language model}
\newacronym[plural={LVLMs}]{lvlm}{LVLM}{Large Visual-language models}
\newacronym{vespa}{VESPA}{Vision-language Enabled Self-supervised Pseudo-label Assignment}

\def\wacvPaperID{*****} 
\def\confName{WACV}
\def\confYear{2026}

\title{VESPA: Towards un(Human)supervised Open-World Pointcloud Labeling for Autonomous Driving}

\author{Levente Tempfli$^*$, Esteban Rivera$^*$, Markus Lienkamp\\
Technical University of Munich\\
Munich, Germany\\
{\tt\small esteban.rivera@tum.de}
}

\begin{document}
\maketitle
\renewcommand{\thefootnote}{\fnsymbol{footnote}}
\footnotetext[1]{Equal contribution. This work is based on Levente's Master Thesis: Vision-language-supervised Pseudo Labeling for 3D Object Detection (Unpublished).}
\begin{abstract}
Data collection for autonomous driving is rapidly accelerating, but manual annotation, especially for 3D labels, remains a major bottleneck due to its high cost and labor intensity. Autolabeling has emerged as a scalable alternative, allowing the generation of labels for point clouds with minimal human intervention. While LiDAR-based autolabeling methods leverage geometric information, they struggle with inherent limitations of lidar data, such as sparsity, occlusions, and incomplete object observations. Furthermore, these methods typically operate in a class-agnostic manner, offering limited semantic granularity. To address these challenges, we introduce VESPA, a multimodal autolabeling pipeline that fuses the geometric precision of LiDAR with the semantic richness of camera images. Our approach leverages vision-language models (VLMs) to enable open-vocabulary object labeling and to refine detection quality directly in the point cloud domain. VESPA supports the discovery of novel categories and produces high-quality 3D pseudolabels without requiring ground-truth annotations or HD maps. On Nuscenes dataset, VESPA achieves an AP of 52.95\% for object discovery and up to 46.54\% for multiclass object detection, demonstrating strong performance in scalable 3D scene understanding. Code will be available upon acceptance.
\end{abstract}

\section{Introduction}
\label{sec:intro}
\begin{figure*}[!htbp]
    \centering
    \includegraphics[width=0.98\linewidth]{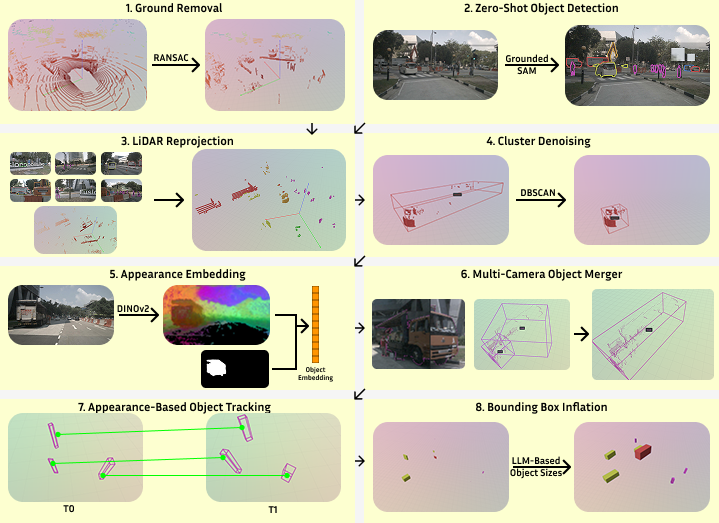}
    \caption{VESPA pipeline overview. Stages: 
    (1) ground removal; (2) zero-shot segmentation; (3) reprojection; (4) cluster denoising; (5) appearance embedding; (6) multi-camera merge; (7) tracking and motion estimation; (8) box refinement.}
    \label{fig:vespa_direct}
\end{figure*}
Accurate and robust 3D perception is fundamental for autonomous driving, providing crucial spatial information (location, dimensions, orientation) essential for path planning, collision avoidance, and predictive behavior modeling~\cite{pendleton2017perception, kocic2018sensors}. Unlike 2D methods, 3D detection offers a comprehensive environmental understanding, directly impacting safety and system efficiency.

Significant advancements in 3D detection have been driven by supervised learning, leveraging large, meticulously labeled datasets of 3D point clouds and images. Deep neural networks are trained to identify and localize objects, demonstrating strong generalization across diverse scenarios~\cite{zhang2025,wangdsvt}. However, this reliance on extensive labeled data introduces substantial challenges. 3D data annotation is exceptionally labor-intensive, time-consuming, and expensive, severely limiting scalability. Consequently, developing strategies to mitigate this heavy dependence on fully labeled data is critical for achieving more adaptable, scalable, and robust 3D detection systems for autonomous driving. 

One promising strategy to reduce the need for manual annotation is autolabeling, which aims to generate 3D bounding box labels without human supervision. These labels can be derived from intrinsic geometric properties of the scene, commonly exploited in the LiDAR domain through self-supervised methods\cite{lentsch2024union,baur2024liso,seidenschwarzsemoli,zhangoyster}, or from pretrained models originally trained on different data distributions. The latter includes both camera-based and LiDAR-based detectors and typically involves some form of domain adaptation to bridge the gap between the source and target domains\cite{tsaims3d}.

While LiDAR-based 3D object detection models consistently outperform camera-only approaches in autonomous driving, primarily due to their direct measurement of 3D spatial information, this advantage holds true largely for well-defined, geometrically distinct object classes \cite{yin2021centernet3d, geiger2012kitti}. However, when distinguishing semantically similar categories (e.g., "car" versus "police car"), the rich semantic features derived from camera images become indispensable for accurate classification \cite{zhou2020centerpoint}. This inherent complementarity logically leads to the necessity of fusing camera and LiDAR data to harness the strengths of both modalities for robust 3D object detection.

Recent years have witnessed a surge in vision-language models (VLMs) and other foundational models, exemplified by architectures like GPT-4~\cite{openai2024gpt4technicalreport}, Qwen-VL~\cite{qwen_vl}, and SAM~\cite{kirillov2023segment}. These models, trained on very large and diverse datasets, have demonstrated remarkable generalization capabilities across a wide spectrum of visual understanding tasks~\cite{peri2023towards}. More critically for autonomous driving, their knowledge base allows them to process and understand a significantly broader range of object classes and edge cases. Leveraging the vast knowledge embedded in these foundational models could be key to enhancing the safety and reliability of autonomous vehicles by improving their ability to recognize infrequent occurrences and novel classes.

In this work, we propose \gls{vespa} to strategically combine the spatial precision of LiDAR with the semantic knowledge of camera information processed by a \gls{vlm}. Our method employs a novel pseudo-labeling pipeline that circumvents the need for human-annotated data, instead relying on the generalized knowledge provided by VLMs and other foundational models. Our contributions are twofold:
\begin{enumerate}
    \item We present \gls{vespa}, a \gls{vlm}-supervised pipeline for point cloud labeling tailored for the autonomous driving application. 
    \item We present a complete characterization of the performance of VESPA with respect to its components both from pointcloud as well as image domain. 
\end{enumerate}


\section{Related work}
\label{sec:related_work}

\subsection{3D LiDAR Autolabeling}
A common strategy for generating labels without human-annotated ground truth leverages geometric cues derived from either single point clouds or sequences of point clouds.

Early approaches operate on individual point cloud frames, using clustering algorithms such as DBSCAN or HDBSCAN~\cite{mcinnes2017hdbscan} to segment object candidates directly. However, due to the limitations of single-frame information—such as occlusions, partial visibility, and sensor noise—these methods often struggle with robustness and completeness.

To address these challenges, several methods exploit temporal information across multiple consecutive frames to improve object discovery. Some extend clustering techniques by incorporating temporal consistency through object tracking~\cite{luo2023reward, zhang2024harnessing}. For example, OYSTER~\cite{zhang2023towards} tracks object clusters over time and enforces geometric consistency across observations. It propagates dense, near-range measurements to augment sparse, far-range detections, thus enhancing label quality over distance. Similarly, MODEST ~\cite{you2022learning} identifies dynamic objects by detecting inconsistencies in point coverage across frames, using these dynamic regions as self-supervised labels. CPD~\cite{wu2024commonsense} further builds upon tracking by introducing object class prototypes, which guide the refinement of bounding boxes according to expected object dimensions.

An even finer integration of temporal cues involves tracking points directly instead of objects. LISO~\cite{baur2024liso}, for instance, estimates point-wise scene flow via Iterative Closest Point (ICP) alignment, and then infers object bounding boxes around coherent motion patterns. SeMoLi~\cite{seidenschwarzsemoli} takes a similar direction by computing point-wise velocities and learning to cluster moving points using Graph Neural Networks (GNNs), under the assumption that points with similar motion likely belong to the same object.

ViLGOD~\cite{fruhwirth2024vision} proposes a novel approach to integrating semantic information into point cloud data by leveraging vision-language models without requiring direct image inputs. Instead of fusing RGB imagery with the point cloud, VILGOD projects detected clusters into depth images, which are then processed by a CLIP model to establish associations with textual labels. This strategy enables the injection of high-level semantic classification signals into the point cloud modality. However, its effectiveness remains constrained by the limited visual fidelity and abstract nature of depth renderings, which can reduce the discriminative power of CLIP in this context.

\subsection{3D Multimodal Autolabeling}
Lidar-only autolabeling approaches leverage rich geometric information, making them highly effective for object localization. However, they typically operate in a class-agnostic manner, which limits their ability to provide semantic labels. This semantic understanding is crucial for downstream modules in the autonomous driving stack, such as prediction and planning. In contrast, camera-based methods can offer detailed classification information, but transferring this knowledge from the image domain to the lidar domain remains a core challenge.

One common strategy to fuse lidar and camera information for autolabeling is to use a lidar-first pipeline, augmenting classless geometric clusters with image-derived semantics. For instance, UNION relies heavily on geometric cues from point clouds to identify object candidates, while incorporating features from foundational vision models to classify objects into static, movable, or moving categories. Despite the integration of image features, the detection backbone remains fundamentally lidar-centric.

Another prominent line of work processes data primarily in the image domain and then transfers semantic information into the 3D space. CM3D\cite{khurana2024shelf}, for example, employs a 2D vision-language model (VLM) to generate open-vocabulary bounding boxes, which are refined into segmentation masks using SAM. These masks are then projected into the point cloud to identify candidate 3D instances, which are further refined using outputs from the VLM. Our method, VESPA, shares similarities with CM3D in leveraging VLMs, but differs fundamentally in architecture and reliance: we prioritize spatial cues from the point cloud over image data, avoid dependence on map information for tracking and velocity estimation, and operate independently of ground-truth labels. Moreover, our approach achieves high-quality autolabels without requiring multiple rounds of self-training.

\section{Method}
\label{sec:method}

We introduce \textbf{VESPA}, a fully unsupervised, vision-language-model-driven pipeline for 3D object discovery and pseudo-label generation, leveraging multi-camera images and LiDAR in automotive-scale environments.
An overview of the pipeline is shown in Figure~\ref{fig:vespa_direct}; VESPA fuses state-of-the-art zero-shot object segmentation, clustering-based geometric denoising, appearance-based tracking, and large-language-model-provided priors to produce high-quality 3D object proposals. The modules of VESPA are described sequentially below.

\subsection{Ground Removal}

Following the approach in UNION~\cite{lentsch2024union}, VESPA performs robust ground and sky removal prior to downstream reprojection. Both global and angular sector-wise ground planes are fitted via RANSAC~\cite{fischler1981random}, using multiple (5) temporally-aligned LiDAR sweeps to improve density. Each point is classified by its signed distance to the local estimated plane and altitude with respect to the vehicle, and points within $30$\,cm of a ground plane are discarded. Points at excessive range ($>40m$) or height ($<4m$) are also filtered.

\subsection{Zero-shot Object Segmentation}

For each camera image, we run Grounding~DINO~\cite{liu2024grounding} with open-vocabulary queries (e.g., \textit{car}, \textit{pedestrian}, \textit{truck}, etc.), resulting in 2D bounding boxes, keeping those with confidence $\geq 0.3$. Segmentation masks for each proposal are generated using Segment Anything Model (SAM)~\cite{kirillov2023segment}. To retain high-quality, unique instances, we employ a custom mask-based NMS: for each proposal, intersection-over-area (IoA) is computed against higher-confidence masks; if $IoA > 0.5$, the proposal is suppressed, otherwise it is trimmed.

\subsection{LiDAR Reprojection and Mask Association}

For each frame, ground-filtered LiDAR points are projected into the image plane of every camera using accurate dataset-provided extrinsic and ego-pose calibration (see~\cite{caesar2020nuscenes} for data conventions). Each 3D point is associated with a segmentation mask if the projection falls inside the mask, resulting in a candidate set of per-object point clusters~\cite{lentsch2024union,khurana2024shelf}.

\subsection{Cluster Denoising via Spatial Clustering}

The association of LiDAR points to camera-based segmentation masks can introduce substantial noise in the resulting object clusters due to inherent differences in sensor perspective, mounting height, and occlusions. In typical automotive setups (e.g., nuScenes~\cite{caesar2020nuscenes}), the LiDAR sensor is mounted significantly higher than the cameras. As a consequence, the LiDAR returns include 3D points from surfaces that are occluded or invisible from the viewpoint of the camera. When these points are projected into the camera image, their 2D positions may still fall within the foreground object mask, even if they lie behind the true intended object.

To robustly filter out these spurious associations, we apply DBSCAN~\cite{ester1996density} clustering to the projected object points in the XY plane. As the distance threshold for clustering, we use the average width of the object class. This value was retrieved from an LLM to avoid dataset-specific bias or manual tuning. DBSCAN, being density-based and agnostic to shape, naturally separates the main object body from disconnected or diffuse background outliers. For each candidate cluster, only the largest spatially connected component is retained, while points lying in sparse or isolated regions are discarded. This denoising procedure greatly improves the reliability and physical plausibility of the resulting 3D object proposals. We subsequently refit the oriented bounding box to the denoised points using L-shape fitting~\cite{zhang2017efficient}.

\subsection{Multi-camera Object Merging}

Objects visible in multiple overlapping camera views are merged by leveraging spatial overlap, shared points, and border-adjacency cues. Two proposals are grouped if their classes match and they either share LiDAR points or are detected near the edges of adjacent camera images—specifically, within the leftmost or rightmost 15\% of the image width. The 3D point clouds, appearance features, and confidence scores for all detections in a group are merged and averaged.

\subsection{Appearance Embedding Extraction}

For each object proposal, we compute an appearance embedding to enable robust tracking and identity association. The segmentation mask in the corresponding camera image is used to extract DINOv2~\cite{oquab2023dinov2} features, which are averaged over the mask area to yield a compact, high-dimensional embedding vector. This embedding captures the object's visual characteristics and category information in a manner robust to moderate viewpoint or shape variations, providing a learned signature for downstream matching.

\subsection{Appearance-Based Tracking and Motion Estimation}

To associate objects across consecutive frames and estimate their motions, we employ a two-step tracking approach. First, each object in the current frame is matched to candidate objects in the next frame by maximizing the cosine similarity of their DINOv2 appearance embeddings, considering only class-matching proposals and applying a spatial proximity threshold based on the maximum velocity of objects in an urban environment, queried from a large language model. Mutual best matches are accepted as identity correspondences.

Once matches are established, we estimate each object’s velocity and precise displacement using rigid alignment: we apply the Iterative Closest Point (ICP) algorithm~\cite{besl1992method} on the object’s LiDAR point clusters to compute the translation vector that best aligns the matched object clouds in consecutive frames. The resulting motion estimate is used for velocity pseudo-labels and for refining bounding box orientation.

\subsection{Bounding Box Refinement and Inflation}

Final object proposals are refined as follows. If an object is detected as moving ($>0.5\,\mathrm{m/s}$), its bounding box is oriented to the estimated direction of motion; otherwise, fallback heuristics based on shape-based priors are invoked. 

Box sizes are compared to dataset-independent class priors sourced from LLMs (\textit{cf.}~\cite{khurana2024shelf}), and excessively small dimensions are inflated to class means. To maintain LiDAR-to-surface alignment, the box is shifted outward, along the axis of inflation, away from the ego vehicle.
\section{Results}

\subsection{Experimental Setup}
\paragraph{Datasets.}We principally evaluate VESPA on the nuScenes dataset, a large-scale urban autonomous driving benchmark comprising 700 training and 150 validation scenes, each equipped with a LiDAR sensor and six cameras providing $360^\circ$ coverage~\cite{caesar2020nuscenes}. All experiments are conducted on these predefined splits, with models trained using pseudo labels generated from the training set and evaluated on ground truth annotations from the validation set. To directly quantify the effect of training, we report pseudo-label quality by comparing pseudo labels directly to ground truth on the validation set, which enables us to measure the improvement resulting from detector training. Additionally, we present results using pseudolabels on a proprietary dataset, here named just AnonymousScenes, which is currently in the process of being made publicly available. This dataset comprises 22 urban driving scenes and a total of 1,960 frames. The sensor configuration is similar to that of nuScenes, but features a denser LiDAR and higher-resolution imagery, with a 1920×1080 front-facing camera and five surrounding cameras operating at 384×240 resolution.

\paragraph{Class Mappings}For object categories, we follow established protocols and focus on 8 dynamic classes of nuScenes (excluding static traffic cone and barrier). To ensure comparability with prior work~\cite{zhang2023towards, baur2024liso, khurana2024shelf, lentsch2024union}, experiments are conducted using three label mappings: 1-class (class-agnostic), 3-class (vehicle, pedestrian, bicycle), and full 8-class setups. The pseudo-label generation pipeline is run per scene with all 8 class names provided as prompts to Grounding DINO~\cite{liu2024grounding}. Class-agnostic (1-class) or 3-class label mappings are achieved by regrouping the resulting labels at the pipeline output stage, without rerunning detection.

\paragraph{Detector}We use OpenPCDet~\cite{openpcdet2020} to train CenterPoint detectors~\cite{yin2021center} on pseudo labels from the training set, adopting default settings of OpenPCDet: a voxel size of $0.075 \times 0.075$ m, batch size of 4, and 20 training epochs. Class-balanced grouping and sampling (CBGS)~\cite{zhu2019class} is enabled for the 3-class and 8-class experiments to mitigate class imbalance; the 1-class setup is trained without class balancing.

\paragraph{Metrics}Performance is assessed with the official nuScenes detection protocol~\cite{caesar2020nuscenes}: mean Average Precision (mAP) is calculated using center distance matching thresholds of 0.5, 1, 2, and 4 meters, alongside true positive errors for translation, scale, orientation, and velocity (mATE, mASE, mAOE, mAVE). The final nuScenes Detection Score (NDS) is reported as a weighted sum of mAP and these error metrics, with attribute errors (mAAE) set to 1.0 as attributes are not predicted.

\paragraph{Tuning} Hyperparameters for the pseudo-label generation pipeline were set by visual tuning on three representative training scenes from nuScenes and then fixed for all scenes in all experiments.

\subsection{Quantitative Results}
We report in ~\Cref{tab:agnostic} the results for class-agnostic detection, where all detected classes are mapped to a single category. This experiment isolates the quality of object discovery, independent of classification. Results are presented for both supervised baselines and prior state-of-the-art unsupervised object discovery methods.

While a notable gap remains between supervised and unsupervised performance, our method, \gls{vespa}, consistently outperforms existing unsupervised approaches. In terms of AP and NDS, \gls{vespa} falls between the performance achieved using 10\% and 5\% of the fully supervised training data. Interestingly, although \gls{vespa}'s TP metrics are slightly below the 5\% supervised baseline, it surpasses it in AOE and AVE.

Compared to prior unsupervised methods on the nuScenes benchmark, \gls{vespa} achieves state-of-the-art performance across all metrics. Specifically, it improves AP by 12.5\% and NDS by 16.9\% relative to UNION.
\begin{table*}[ht]
\centering
\caption{Class-agnostic object detection on the nuScenes validation set after training CenterPoint. \dag We reimplemented UNION based on the original code and present this results here and in the following tables.}
\begin{tabular}{lcccccccc}
\label{tab:agnostic}
Method&Labels&ST&AP$\uparrow$&NDS$\uparrow$&ATE$\downarrow$&ASE$\downarrow$&AOE$\downarrow$&AVE$\downarrow$\\
\midrule
Supervised 1\%&GT&$\times$&31.47&27.78&0.477&0.317&1.568&1.275\\
Supervised 5\%&GT&$\times$&51.79&43.02&0.321&0.248&1.351&0.717\\
Supervised 10\%&GT&$\times$&62.01&50.54&0.286&0.228&0.994&0.537\\
Supervised 100\%&GT&$\times$&82.10&72.26&0.188&0.200&0.274&0.216\\
\midrule
\midrule
OYSTER~\cite{zhang2023towards}&L&\checkmark&9.1&11.5&0.784&0.521&1.514&-\\
LISO~\cite{baur2024liso}&L&\checkmark&10.9&13.9&0.750&\underline{0.409}&1.062&-\\
CMD3D(w/ BEVFusion)~\cite{khurana2024shelf}&L+C&\checkmark&27.9&27.6&0.592&0.428&0.872&-\\
UNION~\dag\cite{lentsch2024union}&L+C&$\times$&\underline{38.4}&\underline{31.2}&\underline{0.589}&0.497&\underline{0.874}&0.836\\
\midrule
\textbf{VESPA}&L+C&$\times$&\textbf{52.95}&\textbf{48.12}&\textbf{0.351}&\textbf{0.350}&\textbf{0.765}&\textbf{0.368}\\
\scriptsize{Standard deviation over 3 runs}&&&\scriptsize{$\pm$0.13}&\scriptsize{$\pm$0.16}&\scriptsize{$\pm$0.0005}&\scriptsize{$\pm$0.0016}&\scriptsize{$\pm$0.0085}&\scriptsize{$\pm$0.0022}\\
\scriptsize{Relative improvement over previous best}&&&\scriptsize{\textcolor{green!50!black}{+37.9\%}}&\scriptsize{\textcolor{green!50!black}{+54.2\%}}&\scriptsize{\textcolor{green!50!black}{-40.4\%}}&\scriptsize{\textcolor{green!50!black}{-14.4\%}}&\scriptsize{\textcolor{green!50!black}{-12.5\%}}&\scriptsize{\textcolor{green!50!black}{-56.0\%}}\\
\bottomrule
\end{tabular}
\end{table*}

\Cref{tab:vespa_vs_union} presents the results for both the 3-class and 8-class configurations of UNION and VESPA, evaluating the quality of the generated pseudolabels by comparing them against a CenterPoint model trained directly on these pseudolabels.

For the 3-class configuration, VESPA achieves a notable improvement in mAP, increasing by 22.2\%, compared to 7.3\% for UNION. Additionally, most TP metrics exhibit reduced error for VESPA, with the exception of AVE. We attribute this to UNION's use of the ICP-Flow approach, which provides more direct velocity estimation, whereas VESPA relies on inferred velocity from its tracking method. Nevertheless, despite the slightly higher AVE, VESPA achieves twice the NDS score of UNION.

In terms of AP across individual classes, VESPA demonstrates significant improvements, particularly for bicycle and pedestrian. Notably, VESPA detects bicycles with an AP of 9\%, while UNION fails to detect them altogether. For pedestrians, VESPA's pseudolabels result in up to a 10× improvement in AP compared to UNION. In contrast, for the car class, UNION slightly outperforms VESPA by approximately 1.2\%. This highlights VESPA's advantage when handling smaller object classes, which benefit from additional image-based features that complement the sparse point cloud information often associated with such objects.

Training a CenterPoint model directly on the generated pseudolabels further improves performance for both methods. Continuing the observed trend, VESPA consistently outperforms UNION across all metrics, with improvements of 19\% in mAP and 18\% in NDS. Remarkably, even for the vehicle class, where UNION previously had an advantage, VESPA now surpasses it with a 10\% AP gain. These results suggest that while the pseudolabels may not be perfect, they are sufficiently high-quality to enable a downstream model to effectively generalize, particularly when trained on broad semantic categories.

For the 8-class configuration, which provides a more fine-grained semantic breakdown of the dataset, the overall metric scores are lower than in the 3-class case. This is expected, as semantically similar classes, such as bicycle and motorcycle, or car, truck, and construction vehicle, are grouped in the 3-class setting, simplifying the task by reducing inter-class ambiguity. Once this inherent difficulty is accounted for, the general performance trend remains consistent with the 3-class case. Notably, VESPA's pseudolabels outperform UNION's by over 3× in mAP and 2× in NDS.

At the class level, VESPA partially detects bicycles, motorcycles, and construction vehicles, which UNION fails to recognize. Even for the vehicle class—where UNION tends to perform well—VESPA achieves slightly better AP, with a 0.7\% improvement. However, trailer remains a challenging category for VESPA, with no detections recorded. When training CenterPoint models on the respective pseudolabels, VESPA continues to outperform UNION across all metrics except for mean Average Orientation Error (mAOE). Crucially, both mAP and NDS improve by more than 2× compared to UNION.

The class-wise improvements are also preserved in the trained CenterPoint models. Categories previously undetected by UNION, such as bicycle, motorcycle, and construction vehicle, now yield non-zero AP values under VESPA. In most cases, the AP of the trained model exceeds that of the original pseudolabels, indicating that the model is able to generalize beyond the initial supervision. The exception is the bicycle class, where the trained model achieves only half the AP of the pseudolabels, possibly due to limited training examples or higher visual ambiguity in this class.

\begin{table}[!htbp]
\centering
\caption[Comparison of VESPA and UNION]{Comparison of VESPA and UNION on the nuScenes dataset across 3-class and 8-class settings. The results include both pseudo label quality on the training set and downstream detection performance on the validation set after training CenterPoint \cite{yin2021center}. For the pseudolabels mAP calculation, all the confidences were set to 1.}
\label{tab:vespa_vs_union}
\begin{tabular}{lcc|cc}
\toprule
\textbf{Metric} & \multicolumn{2}{c|}{\textbf{Pseudo Labels}} & \multicolumn{2}{c}{\textbf{Centerpoint}} \\
& \textbf{UNION} & \textbf{VESPA} & \textbf{UNION} & \textbf{VESPA} \\
\midrule
\multicolumn{5}{c}{\textbf{3-class results}} \\
\midrule
mAP & 7.33 & \textbf{22.22} & 27.19 & \textbf{46.54 } \\
mATE & 0.6629 & \textbf{0.5319} & 0.4280 & \textbf{0.3843 } \\
mASE & 0.7379 & \textbf{0.4180} & 0.5473 & \textbf{0.3649 } \\
mAOE & 1.1122 & \textbf{1.0151} & 1.1463 & \textbf{0.8684 } \\
mAVE & \textbf{0.8649} & 0.8689 & 0.8678 & \textbf{0.3640 } \\
NDS & 11.01 & \textbf{22.95} & 25.17 & \textbf{43.45} \\
\midrule
\textbf{AP} \\
bicycle & 0.000 & \textbf{9.86} & 0.000 & \textbf{15.57 } \\
pedestrian & 3.85 & \textbf{40.04} & 42.87 & \textbf{75.10} \\
vehicle & \textbf{18.14} & 16.91 & 38.71 & \textbf{48.96} \\
\midrule
\multicolumn{5}{c}{\textbf{8-class results}} \\
\midrule
mAP & 3.32 & \textbf{10.96} & 11.98 & \textbf{24.82} \\
mATE & 0.7928 & \textbf{0.7397} & 0.6844 & \textbf{0.5995 } \\
mASE & 0.7686 & \textbf{0.4998} & 0.6812 & \textbf{0.3746 } \\
mAOE & 1.0882 & \textbf{1.0677} & \textbf{1.0765} & 1.0788\\
mAVE & 1.0122 & \textbf{0.8391} & 0.9301 & \textbf{0.3666} \\
NDS & 6.05 & \textbf{14.70} & 13.03 & \textbf{29.00} \\
\midrule
\textbf{AP} \\
bicycle & 0.000 & \textbf{8.11} & 0.000 & \textbf{4.36} \\
bus & 4.71 & \textbf{7.36} & 8.26 & \textbf{27.64} \\
car & 16.61 & \textbf{17.27} & 38.78 & \textbf{52.19} \\
cons. vehicle & 0.000 & \textbf{2.37} & 0.000 & \textbf{2.28 } \\
motorcycle & 0.000 & \textbf{8.46} & 0.000 & \textbf{20.05 } \\
pedestrian & 3.85 & \textbf{40.04} & 43.77 & \textbf{74.00} \\
trailer & 0.000 & 0.000 & 0.000 & 0.000 \\
truck & 1.39 & \textbf{4.11} & 5.71 & \textbf{18.02} \\
\bottomrule
\end{tabular}
\end{table}

\paragraph{AnonymousScenes}
In \Cref{tab:edgar_vs_nuscenes}, we report the pseudolabeling performance of VESPA evaluated on the \textit{AnonymousScenes} dataset. Overall, the metrics are noticeably lower compared to those obtained on nuScenes. To contextualize this performance drop, it is important to consider the differences between the two datasets.

AnonymousScenes employs a denser LiDAR sensor with 64 channels, in contrast to the 32-channel sensor used in nuScenes. However, all surrounding cameras in AnonymousScenes—except for the front-facing one—have a significantly lower resolution of 380×240 pixels. The results suggest that VESPA is relatively robust to variations in LiDAR configuration, but is highly sensitive to image resolution. In particular, low-resolution imagery appears to substantially degrade the quality of the VLM-based detections, likely due to the inability to resolve small or fine-grained objects. This highlights a limitation of current VLMs when applied to low-resolution inputs and emphasizes the importance of high-quality visual data for the success of the VESPA pipeline.

\begin{table}[!htbp]
\centering
\caption[Evaluation of the pseudolabels in AnonymousScenes]{Evaluation of the pseudolabels in AnonymousScenes}
\label{tab:edgar_vs_nuscenes}
\begin{tabular}{lc}
\toprule
\textbf{Metric} & \textbf{AnonymousScenes} \\
mAP  & 4.72 \\
mATE  & 0.8013  \\
mASE  & 0.5487 \\
mAOE  & 1.4734 \\
mAVE & 5.9457  \\
NDS  & 8.86 \\
\midrule
\textbf{Per-Class AP} \\
bicycle & 1.46 \\
bus  & 5.86 \\
car  & 15.14  \\
construction vehicle  & 0.000  \\
motorcycle  & 2.11 \\
pedestrian  & 10.81  \\
trailer & 0.000  \\
truck & 2.40 \\
\bottomrule
\end{tabular}%
\end{table}
\subsection{Ablation Studies}

\paragraph{Pointcloud ablation}\Cref{tab:ablation} presents the ablation study for the VESPA pipeline. The full configuration achieves the best overall performance across all metrics, with the exception of a minor 2\% decrease in AP for the bicycle class. Among all components, the tracking module (TR) has the most substantial impact, with its removal resulting in a 21\% drop in mAP and a 22\% drop in NDS. This effect is expected, as the TR module addresses one of the primary challenges in 3D scene understanding: incomplete or occluded measurements. By leveraging temporal information, the tracking module stabilizes detections over time, providing more coherent object trajectories and contributing to improvements in metrics such as Average Orientation Error (AOE).

We also observe that the cluster denoising module (DN) has a pronounced influence on mAP, with a degradation of up to 19\% when omitted. This is attributed to the sensitivity of the NuScenes evaluation protocol to bounding box localization—particularly the center position. Noisy clusters can severely distort box centers, leading to mismatched predictions and lower AP scores. Since box translation, rotation, and size also contribute to the NDS metric, their degradation in the presence of noise further explains the performance drop. Interestingly, a configuration that includes only DN, while omitting the other modules, outperforms the version that uses all modules except DN, highlighting its central role in ensuring overall robustness.

From these results, it becomes evident that DN plays a critical role in robustifying the system, first, by mitigating calibration and synchronization errors between the camera and LiDAR, which can cause inconsistencies at object boundaries; and second, by addressing parallax-related distortions, which occur even under ideal calibration due to viewpoint differences between sensors.

In contrast, the multi-camera matching (MCM) and box inflation (BI) modules provide more incremental benefits. Removing MCM results in a 1.2\% drop, and removing BI leads to a 6\% decrease in performance. These modules contribute to smoother spatial consistency across views and improved box dimensions, which refine, but do not fundamentally alter, the system’s detection capabilities.

Finally, the last row of the table simulates a scenario with ideal sensing conditions: Perfect calibration, no occlusions, complete detections, and no inter-camera overlap, eliminating the need for corrective modules. Interestingly, this configuration yields the worst performance, reaffirming that VESPA's strength lies precisely in its robust correction modules, which effectively handle the real-world imperfections often ignored by more naïve VLM-to-pointcloud distillation methods such as CM3D\cite{khurana2024shelf}.

\begin{table*}[ht]
\centering
\caption{Ablation of 3-class object detection with VESPA on the nuScenes validation set after training CenterPoint. First 4 columns represent the modules: DN=Cluster denoising; MCM=Multi-camera object merger; TR=Tracking and yaw adjustment; BI=Bounding box inflation}
\begin{tabular}{lcccccccccccc}
\toprule
DN&MCM&TR&BI&mAP$\uparrow$&NDS$\uparrow$&ATE$\downarrow$&ASE$\downarrow$&AOE$\downarrow$&AVE$\downarrow$&$\textrm{AP}_{\textrm{car}}\uparrow$&$\textrm{AP}_{\textrm{pedestrian}}\uparrow$&$\textrm{AP}_{\textrm{cyclist}}\uparrow$\\
\midrule     
\checkmark&\checkmark&\checkmark&\checkmark&\textbf{46.76}&\textbf{43.47}&\textbf{0.384}&\textbf{0.363}&\textbf{0.877}&\textbf{0.365}&\textbf{48.9}&\textbf{75.2}&16.1\\
$\times$&\checkmark&\checkmark&\checkmark&27.86&26.72&0.593&0.442&0.982&0.702&23.8&46.4&13.4\\
\checkmark&$\times$&\checkmark&\checkmark&\underline{45.50}&\underline{42.44}&0.396&\underline{0.364}&\underline{0.887}&\underline{0.382}&\underline{45.6}&\textbf{75.2}&15.7\\
\checkmark&\checkmark&$\times$&\checkmark&25.48&21.84&0.658&0.431&1.483&1.720&18.7&45.6&12.2\\
\checkmark&\checkmark&\checkmark&$\times$&40.68&35.71&0.400&0.766&0.902&0.393&30.9&\underline{73.8}&\underline{17.3}\\
\checkmark&\checkmark&$\times$&$\times$&40.35&28.48&\underline{0.393}&0.776&1.473&1.650&30.4&72.6&\textbf{18.1}\\
\checkmark&$\times$&$\times$&$\times$&38.69&27.20&0.422&0.792&1.459&1.630&25.9&73.1&17.2\\
$\times$&$\times$&$\times$&$\times$&23.55&16.93&0.655&0.828&1.467&1.716&14.4&43.7&12.5\\
\bottomrule
\label{tab:ablation}
\end{tabular}
\end{table*}

\paragraph{Image ablation}

The image circuit of VESPA is inherently dependent on the quality of the VLM, as it directly determines which objects are detected and subsequently projected onto the point cloud. To quantify this dependency, we compare the performance of the pipeline using two different VLMs: GroundingDINO, which has been used throughout the previous experiments, and OWLv2\cite{minderer2023scaling}, a recent VLM with competitive results in open-world object detection. The results of this comparison are presented in \Cref{tab:vlm_vs_google}.

Although OWLv2 provides a slight improvement across most metrics, these gains are relatively minor and class-dependent. For instance, GroundingDINO performs better for the pedestrian and construction vehicle categories. These results highlight that the VLM introduces an upper bound on the pipeline's performance in terms of object classification and discovery.

Importantly, the modular design of VESPA allows for seamless replacement or upgrading of the VLM without requiring any structural changes to the rest of the pipeline. As VLMs continue to improve, the overall performance of the system can benefit directly from advancements in vision-language modeling.

\begin{table}[h!]
\centering
\caption{mAP comparison between VESPA with GroundingDINO and VESPA with OWLv2 when training a Centerpoint model}
\begin{tabular}{lcc}
\hline
\textbf{Metric} & \textbf{VESPA } & \textbf{VESPA} \\
&Grounding DINO & OWLv2 \\
\hline
mAP & 24.82 & \textbf{26.04} \\
NDS & 29.00 & \textbf{29.18} \\
\midrule
Car AP & 52.20 & \textbf{52.40} \\
Truck AP & 18.00 & \textbf{20.10} \\
Bus AP & 27.60 & \textbf{32.90} \\
Trailer AP & 0.00 & \textbf{0.50} \\
Constr. Vehicle AP & \textbf{2.30} & 1.10 \\
Pedestrian AP & \textbf{74.00} & 68.50 \\
Motorcycle AP & 20.10 & \textbf{26.30} \\
Bicycle AP & 4.40 & \textbf{6.60} \\
\hline
\end{tabular}

\label{tab:vlm_vs_google}
\end{table}
\vspace{-0.2cm}

\section{Discussion}
Overall, VESPA demonstrates strong performance in both class-agnostic object discovery and multiclass detection, achieving consistent and substantial improvements across all evaluation metrics compared to state-of-the-art methods.

A central question arises regarding the fairness of comparing VESPA with methods such as UNION or other unsupervised object discovery and autolabeling approaches. While VESPA does not rely on human-annotated ground truth labels, it does leverage supervision in the form of vision-language model (VLM) outputs, effectively transferring knowledge distilled from the camera domain. In this sense, VESPA is not entirely unsupervised, and its most appropriate comparison may be with methods like CM3D, which also utilize VLM-based supervision.

Nevertheless, if the objective is to reduce or eliminate manual supervision, VESPA marks a significant step forward. With minimal post-processing, it efficiently leverages VLM-derived semantic information to enhance the geometric understanding from point clouds. Even though our experiments use standard NuScenes classes for evaluation, the classification capabilities of VESPA are inherently tied to the VLM employed. As a result, the pseudolabels produced are as open-world as the underlying VLM allows. Unlike UNION’s soft supervision in a fixed label space, VESPA supports open-vocabulary autolabeling by design.

Despite these advantages, a notable performance gap remains compared to fully supervised models, even those trained with just 10\% of labeled data. This highlights that while VLM-based supervision significantly narrows the gap, there is substantial room for improvement. Continued progress in VLM architectures is likely to further boost the performance of methods like VESPA.

Finally, our experiments on the AnonymousScenes dataset reveal a strong sensitivity to image resolution, which directly impacts VLM label quality. In contrast, increasing LiDAR resolution did not yield noticeable performance gains, suggesting that future improvements should prioritize enhancing image-level understanding.

\section{Conclusion}

We introduced VESPA, a VLM-supervised labeling pipeline designed to generate open-world labels for 3D object detection from LiDAR data without human supervision. Building on existing concepts such as image-to-point cloud distillation, VLM-guided supervision, and appearance-based embeddings, we refine and integrate these components into a unified framework. To the best of our knowledge, VESPA achieves state-of-the-art AP for object discovery in the context of un(human)-supervised object discovery and detection.
{
    \small
    \bibliographystyle{ieeenat_fullname}
    \bibliography{main}

\begin{thebibliography}{35}
\providecommand{\natexlab}[1]{#1}
\providecommand{\url}[1]{\texttt{#1}}
\expandafter\ifx\csname urlstyle\endcsname\relax
  \providecommand{\doi}[1]{doi: #1}\else
  \providecommand{\doi}{doi: \begingroup \urlstyle{rm}\Url}\fi

\bibitem[Bai et~al.(2023)Bai, Tang, Long, Fang, Li, Zhang, Wang, Yang, Ding,
  Ma, et~al.]{qwen_vl}
Jinze Bai, Shuai Tang, Fan Long, Jianwei Fang, Yang Li, Bo Zhang, Wenyuan Wang,
  Junyi Yang, Shujie Ding, Xin Ma, et~al.
\newblock Qwen-vl: A strong large vision-language model for practicable
  applications.
\newblock \emph{arXiv preprint arXiv:2308.12966}, 2023.

\bibitem[Baur et~al.(2024)Baur, Moosmann, and Geiger]{baur2024liso}
Stefan~Andreas Baur, Frank Moosmann, and Andreas Geiger.
\newblock Liso: Lidar-only self-supervised 3d object detection.
\newblock In \emph{European Conference on Computer Vision}, pages 253--270.
  Springer, 2024.

\bibitem[Besl and McKay(1992)]{besl1992method}
Paul~J Besl and Neil~D McKay.
\newblock Method for registration of 3-d shapes.
\newblock In \emph{Sensor fusion IV: control paradigms and data structures},
  pages 586--606. Spie, 1992.

\bibitem[Caesar et~al.(2020)Caesar, Bankiti, Lang, Vora, Liong, Xu, Krishnan,
  Pan, Baldan, and Beijbom]{caesar2020nuscenes}
Holger Caesar, Varun Bankiti, Alex~H Lang, Sourabh Vora, Venice~Erin Liong,
  Qiang Xu, Anush Krishnan, Yu Pan, Giancarlo Baldan, and Oscar Beijbom.
\newblock nuscenes: A multimodal dataset for autonomous driving.
\newblock In \emph{Proceedings of the IEEE/CVF conference on computer vision
  and pattern recognition}, pages 11621--11631, 2020.

\bibitem[Ester et~al.(1996)Ester, Kriegel, Sander, Xu,
  et~al.]{ester1996density}
Martin Ester, Hans-Peter Kriegel, J{\"o}rg Sander, Xiaowei Xu, et~al.
\newblock A density-based algorithm for discovering clusters in large spatial
  databases with noise.
\newblock In \emph{kdd}, pages 226--231, 1996.

\bibitem[Fischler and Bolles(1981)]{fischler1981random}
Martin~A Fischler and Robert~C Bolles.
\newblock Random sample consensus: a paradigm for model fitting with
  applications to image analysis and automated cartography.
\newblock \emph{Communications of the ACM}, 24\penalty0 (6):\penalty0 381--395,
  1981.

\bibitem[Fruhwirth-Reisinger et~al.(2024)Fruhwirth-Reisinger, Lin, Mali{\'c},
  Bischof, and Possegger]{fruhwirth2024vision}
Christian Fruhwirth-Reisinger, Wei Lin, Du{\v{s}}an Mali{\'c}, Horst Bischof,
  and Horst Possegger.
\newblock Vision-language guidance for lidar-based unsupervised 3d object
  detection.
\newblock \emph{arXiv preprint arXiv:2408.03790}, 2024.

\bibitem[Geiger et~al.(2012)Geiger, Lenz, and Urtasun]{geiger2012kitti}
Andreas Geiger, Philipp Lenz, and Raquel Urtasun.
\newblock Are we ready for autonomous driving? the kitti vision benchmark
  suite.
\newblock In \emph{2012 IEEE conference on computer vision and pattern
  recognition}, pages 3354--3361. IEEE, 2012.

\bibitem[Khurana et~al.(2024)Khurana, Peri, Hays, and
  Ramanan]{khurana2024shelf}
Mehar Khurana, Neehar Peri, James Hays, and Deva Ramanan.
\newblock Shelf-supervised cross-modal pre-training for 3d object detection.
\newblock \emph{arXiv preprint arXiv:2406.10115}, 2024.

\bibitem[Kirillov et~al.(2023)Kirillov, Mintun, Ravi, Mao, Rolland, Gustafson,
  Xiao, Whitehead, Berg, Lo, et~al.]{kirillov2023segment}
Alexander Kirillov, Eric Mintun, Nikhila Ravi, Hanzi Mao, Chloe Rolland, Laura
  Gustafson, Tete Xiao, Spencer Whitehead, Alexander~C Berg, Wan-Yen Lo, et~al.
\newblock Segment anything.
\newblock In \emph{Proceedings of the IEEE/CVF international conference on
  computer vision}, pages 4015--4026, 2023.

\bibitem[Koci{\'c} et~al.(2018)Koci{\'c}, Jovi{\v{c}}i{\'c}, and
  Drndarevi{\'c}]{kocic2018sensors}
Jelena Koci{\'c}, Nenad Jovi{\v{c}}i{\'c}, and Vujo Drndarevi{\'c}.
\newblock Sensors and sensor fusion in autonomous vehicles.
\newblock In \emph{2018 26th Telecommunications Forum (TELFOR)}, pages
  420--425. IEEE, 2018.

\bibitem[Lentsch et~al.(2024)Lentsch, Caesar, and Gavrila]{lentsch2024union}
Ted Lentsch, Holger Caesar, and Dariu Gavrila.
\newblock Union: Unsupervised 3d object detection using object appearance-based
  pseudo-classes.
\newblock \emph{Advances in Neural Information Processing Systems},
  37:\penalty0 22028--22046, 2024.

\bibitem[Liu et~al.(2024)Liu, Zeng, Ren, Li, Zhang, Yang, Jiang, Li, Yang, Su,
  et~al.]{liu2024grounding}
Shilong Liu, Zhaoyang Zeng, Tianhe Ren, Feng Li, Hao Zhang, Jie Yang, Qing
  Jiang, Chunyuan Li, Jianwei Yang, Hang Su, et~al.
\newblock Grounding dino: Marrying dino with grounded pre-training for open-set
  object detection.
\newblock In \emph{European Conference on Computer Vision}, pages 38--55.
  Springer, 2024.

\bibitem[Luo et~al.(2023)Luo, Liu, Chen, You, Benaim, Phoo, Campbell, Sun,
  Hariharan, and Weinberger]{luo2023reward}
Katie Luo, Zhenzhen Liu, Xiangyu Chen, Yurong You, Sagie Benaim, Cheng~Perng
  Phoo, Mark Campbell, Wen Sun, Bharath Hariharan, and Kilian~Q Weinberger.
\newblock Reward finetuning for faster and more accurate unsupervised object
  discovery.
\newblock \emph{Advances in Neural Information Processing Systems},
  36:\penalty0 13250--13266, 2023.

\bibitem[McInnes et~al.(2017)McInnes, Healy, Astels,
  et~al.]{mcinnes2017hdbscan}
Leland McInnes, John Healy, Steve Astels, et~al.
\newblock hdbscan: Hierarchical density based clustering.
\newblock \emph{J. Open Source Softw.}, 2\penalty0 (11):\penalty0 205, 2017.

\bibitem[Minderer et~al.(2023)Minderer, Gritsenko, and
  Houlsby]{minderer2023scaling}
Matthias Minderer, Alexey Gritsenko, and Neil Houlsby.
\newblock Scaling open-vocabulary object detection.
\newblock \emph{Advances in Neural Information Processing Systems},
  36:\penalty0 72983--73007, 2023.

\bibitem[OpenAI et~al.(2024)OpenAI, Zhang, Zhao, Zheng, Zhuang, Zhuk, and
  Zoph]{openai2024gpt4technicalreport}
OpenAI, Josh Achiam~Marvin Zhang, Shengjia Zhao, Tianhao Zheng, Juntang Zhuang,
  William Zhuk, and Barret Zoph.
\newblock Gpt-4 technical report, 2024.

\bibitem[Oquab et~al.(2023)Oquab, Darcet, Moutakanni, Vo, Szafraniec, Khalidov,
  Fernandez, Haziza, Massa, El-Nouby, et~al.]{oquab2023dinov2}
Maxime Oquab, Timoth{\'e}e Darcet, Th{\'e}o Moutakanni, Huy Vo, Marc
  Szafraniec, Vasil Khalidov, Pierre Fernandez, Daniel Haziza, Francisco Massa,
  Alaaeldin El-Nouby, et~al.
\newblock Dinov2: Learning robust visual features without supervision.
\newblock \emph{arXiv preprint arXiv:2304.07193}, 2023.

\bibitem[Pendleton et~al.(2017)Pendleton, Andersen, Du, Shen, Meghjani, Eng,
  Rus, and Ang]{pendleton2017perception}
Scott~Drew Pendleton, Hans Andersen, Xinxin Du, Xiaotong Shen, Malika Meghjani,
  You~Hong Eng, Daniela Rus, and Marcelo~H Ang.
\newblock Perception, planning, control, and coordination for autonomous
  vehicles.
\newblock \emph{Machines}, 5\penalty0 (1):\penalty0 6, 2017.

\bibitem[Peri et~al.(2023)Peri, Li, Tang, Chen, Li, Wu, Yu, and
  Yang]{peri2023towards}
Karthik Peri, Kuang Li, Jianing Tang, Xinyu Chen, Xuan Li, Xiaokang Wu, Xinge
  Yu, and Bo Yang.
\newblock Towards long-tailed 3d detection.
\newblock In \emph{Proceedings of the IEEE/CVF Conference on Computer Vision
  and Pattern Recognition (CVPR) Workshops}, pages 5556--5566, 2023.

\bibitem[Seidenschwarz et~al.(2024)Seidenschwarz, Osep, Ferroni, Lucey, and
  Leal-Taixe]{seidenschwarzsemoli}
Jenny Seidenschwarz, Aljosa Osep, Franceso Ferroni, Simon Lucey, and Laura
  Leal-Taixe.
\newblock { SeMoLi: What Moves Together Belongs Together }.
\newblock In \emph{2024 IEEE/CVF Conference on Computer Vision and Pattern
  Recognition (CVPR)}, pages 14685--14694, Los Alamitos, CA, USA, 2024. IEEE
  Computer Society.

\bibitem[Team(2020)]{openpcdet2020}
OpenPCDet~Development Team.
\newblock Openpcdet: An open-source toolbox for 3d object detection from point
  clouds.
\newblock \url{https://github.com/open-mmlab/OpenPCDet}, 2020.

\bibitem[Tsai et~al.(2024)Tsai, Berrio, Shan, Nebot, and Worrall]{tsaims3d}
Darren Tsai, Julie~Stephany Berrio, Mao Shan, Eduardo Nebot, and Stewart
  Worrall.
\newblock Ms3d++: Ensemble of experts for multi-source unsupervised domain
  adaptation in 3d object detection.
\newblock \emph{IEEE Transactions on Intelligent Vehicles}, pages 1--16, 2024.

\bibitem[Wang et~al.(2023)Wang, Shi, Shi, Lei, Wang, He, Schiele, and
  Wang]{wangdsvt}
Haiyang Wang, Chen Shi, Shaoshuai Shi, Meng Lei, Sen Wang, Di He, Bernt
  Schiele, and Liwei Wang.
\newblock Dsvt: Dynamic sparse voxel transformer with rotated sets.
\newblock In \emph{2023 IEEE/CVF Conference on Computer Vision and Pattern
  Recognition (CVPR)}, pages 13520--13529, 2023.

\bibitem[Wu et~al.(2024)Wu, Zhao, Huang, Wen, Li, and Wang]{wu2024commonsense}
Hai Wu, Shijia Zhao, Xun Huang, Chenglu Wen, Xin Li, and Cheng Wang.
\newblock Commonsense prototype for outdoor unsupervised 3d object detection.
\newblock In \emph{Proceedings of the IEEE/CVF Conference on Computer Vision
  and Pattern Recognition}, pages 14968--14977, 2024.

\bibitem[Yin et~al.(2021{\natexlab{a}})Yin, Zhou, and
  Krahenbuhl]{yin2021center}
Tianwei Yin, Xingyi Zhou, and Philipp Krahenbuhl.
\newblock Center-based 3d object detection and tracking.
\newblock In \emph{Proceedings of the IEEE/CVF conference on computer vision
  and pattern recognition}, pages 11784--11793, 2021{\natexlab{a}}.

\bibitem[Yin et~al.(2021{\natexlab{b}})Yin, Shi, and Sun]{yin2021centernet3d}
Zheng Yin, Jianping Shi, and Xiaojuan Sun.
\newblock Centernet3d: An anchor-free approach for scalable 3d object
  detection.
\newblock \emph{arXiv preprint arXiv:2104.09503}, 2021{\natexlab{b}}.

\bibitem[You et~al.(2022)You, Luo, Phoo, Chao, Sun, Hariharan, Campbell, and
  Weinberger]{you2022learning}
Yurong You, Katie Luo, Cheng~Perng Phoo, Wei-Lun Chao, Wen Sun, Bharath
  Hariharan, Mark Campbell, and Kilian~Q Weinberger.
\newblock Learning to detect mobile objects from lidar scans without labels.
\newblock In \emph{Proceedings of the IEEE/CVF Conference on Computer Vision
  and Pattern Recognition}, pages 1130--1140, 2022.

\bibitem[Zhang et~al.(2023{\natexlab{a}})Zhang, Yang, Xiong, Casas, Yang, Ren,
  and Urtasun]{zhang2023towards}
Lunjun Zhang, Anqi~Joyce Yang, Yuwen Xiong, Sergio Casas, Bin Yang, Mengye Ren,
  and Raquel Urtasun.
\newblock Towards unsupervised object detection from lidar point clouds.
\newblock In \emph{Proceedings of the IEEE/CVF Conference on Computer Vision
  and Pattern Recognition}, pages 9317--9328, 2023{\natexlab{a}}.

\bibitem[Zhang et~al.(2023{\natexlab{b}})Zhang, Yang, Xiong, Casas, Yang, Ren,
  and Urtasun]{zhangoyster}
Lunjun Zhang, Anqi~Joyce Yang, Yuwen Xiong, Sergio Casas, Bin Yang, Mengye Ren,
  and Raquel Urtasun.
\newblock { Towards Unsupervised Object Detection from LiDAR Point Clouds }.
\newblock In \emph{2023 IEEE/CVF Conference on Computer Vision and Pattern
  Recognition (CVPR)}, pages 9317--9328, Los Alamitos, CA, USA,
  2023{\natexlab{b}}. IEEE Computer Society.

\bibitem[Zhang et~al.(2024)Zhang, Zhang, Yu, and Zheng]{zhang2024harnessing}
Ruiyang Zhang, Hu Zhang, Hang Yu, and Zhedong Zheng.
\newblock Harnessing uncertainty-aware bounding boxes for unsupervised 3d
  object detection.
\newblock \emph{arXiv preprint arXiv:2408.00619}, 2024.

\bibitem[Zhang et~al.(2017)Zhang, Xu, Dong, and Dolan]{zhang2017efficient}
Xiao Zhang, Wenda Xu, Chiyu Dong, and John~M Dolan.
\newblock Efficient l-shape fitting for vehicle detection using laser scanners.
\newblock In \emph{2017 IEEE Intelligent Vehicles Symposium (IV)}, pages
  54--59. IEEE, 2017.

\bibitem[Zhang et~al.(2025)Zhang, Wang, and Dong]{zhang2025}
Xiang Zhang, Hai Wang, and Haoran Dong.
\newblock A survey of deep learning-driven 3d object detection: Sensor
  modalities, technical architectures, and applications.
\newblock \emph{Sensors}, 25\penalty0 (12), 2025.

\bibitem[Zhou et~al.(2020)Zhou, Wang, and
  Kr{\"a}henb{\"u}hl]{zhou2020centerpoint}
Xin Zhou, Daquan Wang, and Philipp Kr{\"a}henb{\"u}hl.
\newblock Centerpoint: Keypoint head for 3d object detection.
\newblock In \emph{Proceedings of the IEEE/CVF conference on computer vision
  and pattern recognition}, pages 12693--12702, 2020.

\bibitem[Zhu et~al.(2019)Zhu, Jiang, Zhou, Li, and Yu]{zhu2019class}
Benjin Zhu, Zhengkai Jiang, Xiangxin Zhou, Zeming Li, and Gang Yu.
\newblock Class-balanced grouping and sampling for point cloud 3d object
  detection.
\newblock \emph{arXiv preprint arXiv:1908.09492}, 2019.

\end{thebibliography}
}
\newpage
\onecolumn
\section{Qualitative examples}
In this section, we qualitatively compare the pseudo-labels generated by UNION \cite{lentsch2024union}, the current state-of-the-art method, with those from our proposed approach, \gls{vespa}. We examine a set of randomly selected frames and analyze the 3D bounding boxes projected to the bird’s eye view. Our focus is on differences in detection quality, particularly the handling of small and distant objects, as well as bounding box accuracy (including size and orientation) and the rate of false positives.

\begin{figure}[h]
    \centering
    \includegraphics[width=0.48\linewidth]{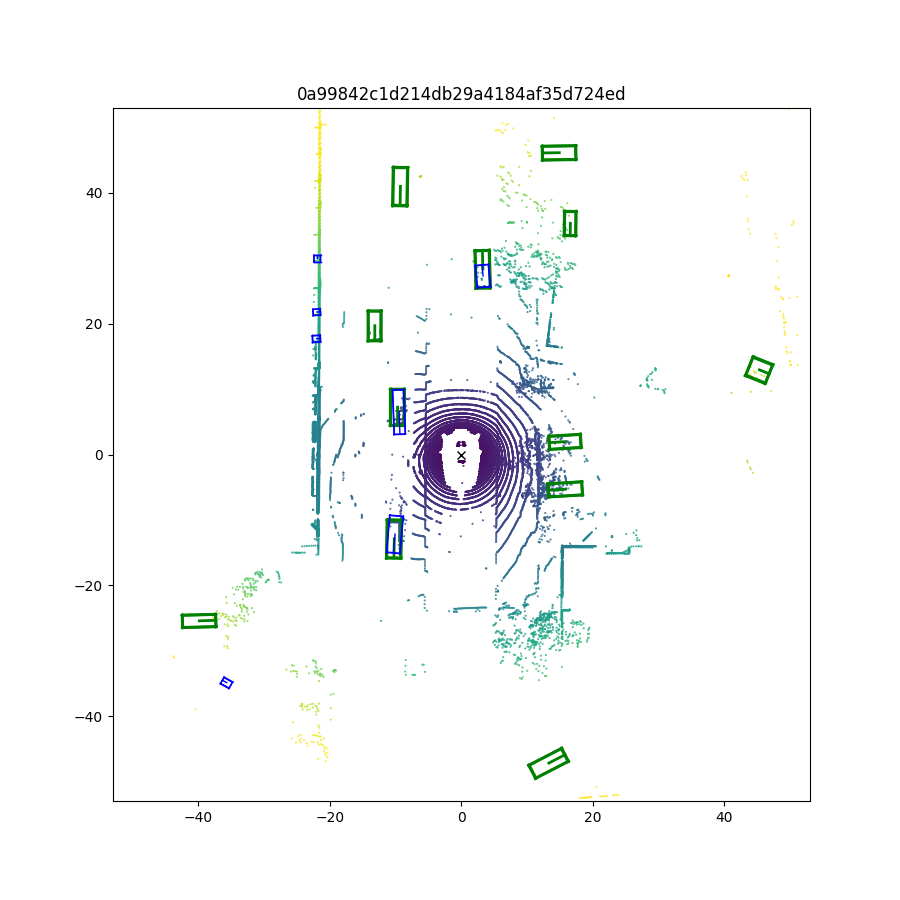}
    \includegraphics[width=0.48\linewidth]{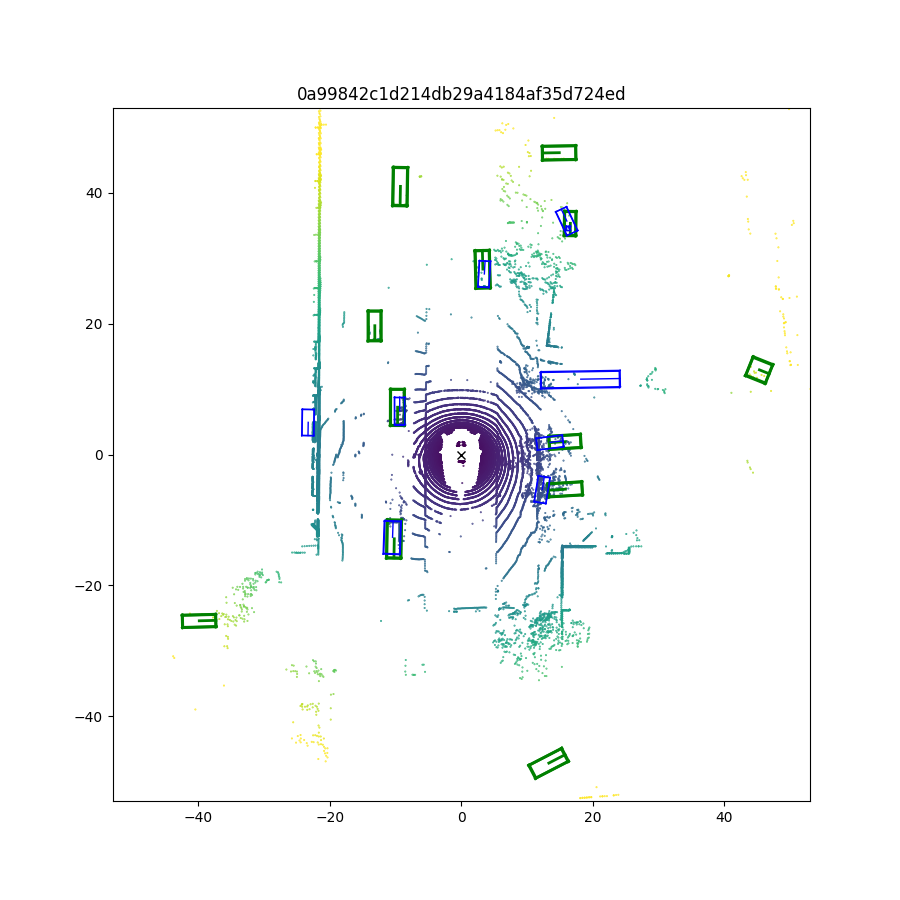}
    \caption[Qualitative comparison between UNION and  VESPA-Direct 1]{Qualitative comparison between UNION (left) and  \gls{vespa}-Direct (right). In the first scene (Figure~\ref{fig:qual_1}), UNION produces a few false positives corresponding to smaller objects and does not estimate the sizes of the true positives as accurately. \gls{vespa}-Direct also introduces two false positives, this time corresponding to larger objects, but provides more complete and correctly sized detections of the dominant objects in the scene.}
    \label{fig:qual_1}
\end{figure}

\begin{figure}[h]
    \centering
    \includegraphics[width=0.48\linewidth]{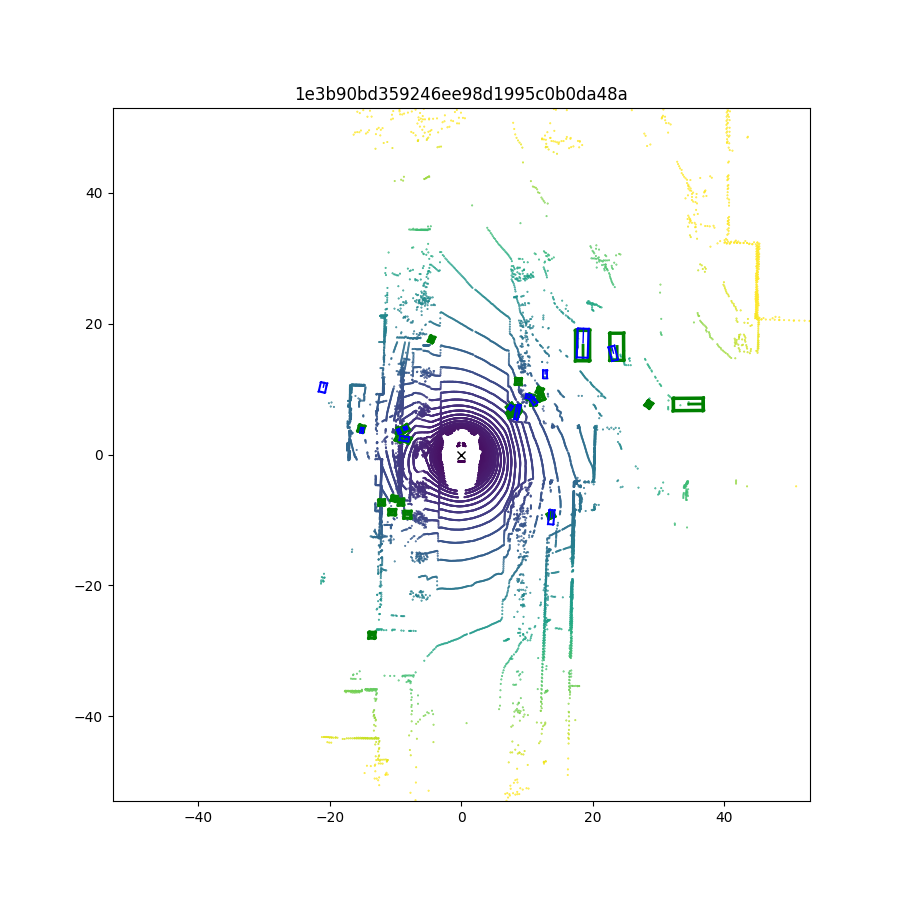}
    \includegraphics[width=0.48\linewidth]{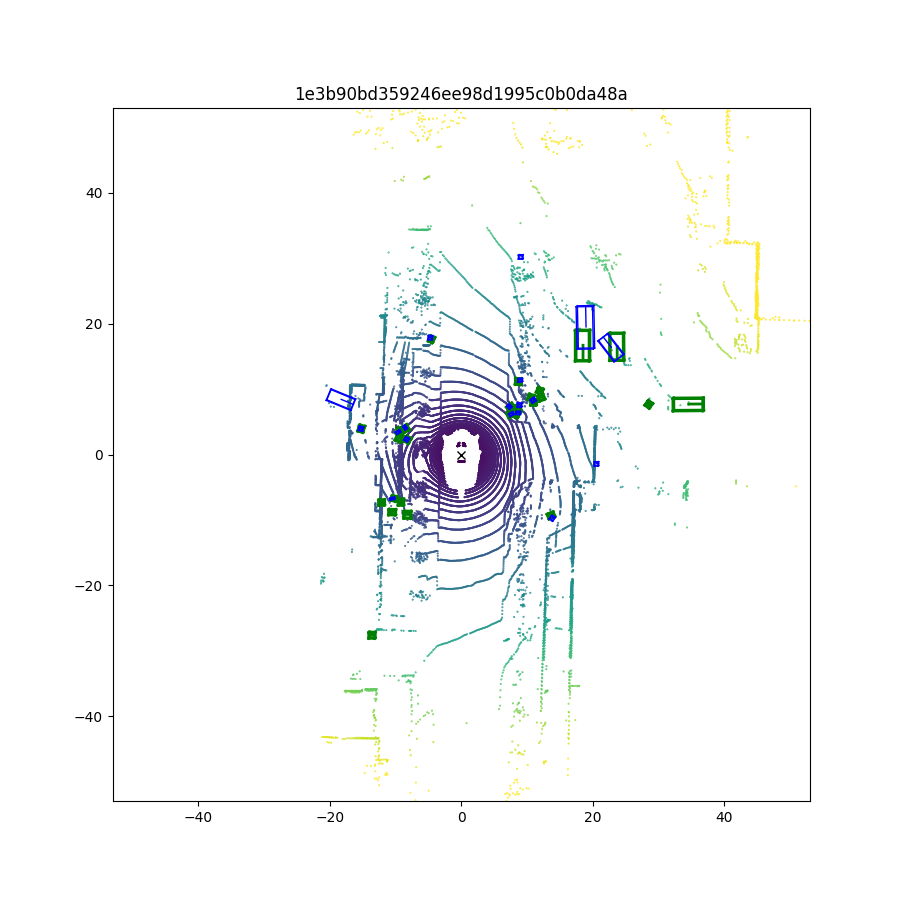}
    \caption[Qualitative comparison between UNION and  VESPA-Direct 2]{Qualitative comparison between UNION (left) and  \gls{vespa}-Direct (right). In the example shown in Figure~\ref{fig:qual_2},  \gls{vespa}-Direct shows a clear advantage in separating and detecting smaller objects. UNION detects one large object more accurately, but misestimates another.  \gls{vespa}-Direct’s prediction for that object is slightly tilted, though overall object coverage is higher.}
    \label{fig:qual_2}
\end{figure}

\begin{figure}[h]
    \centering
    \includegraphics[width=0.48\linewidth]{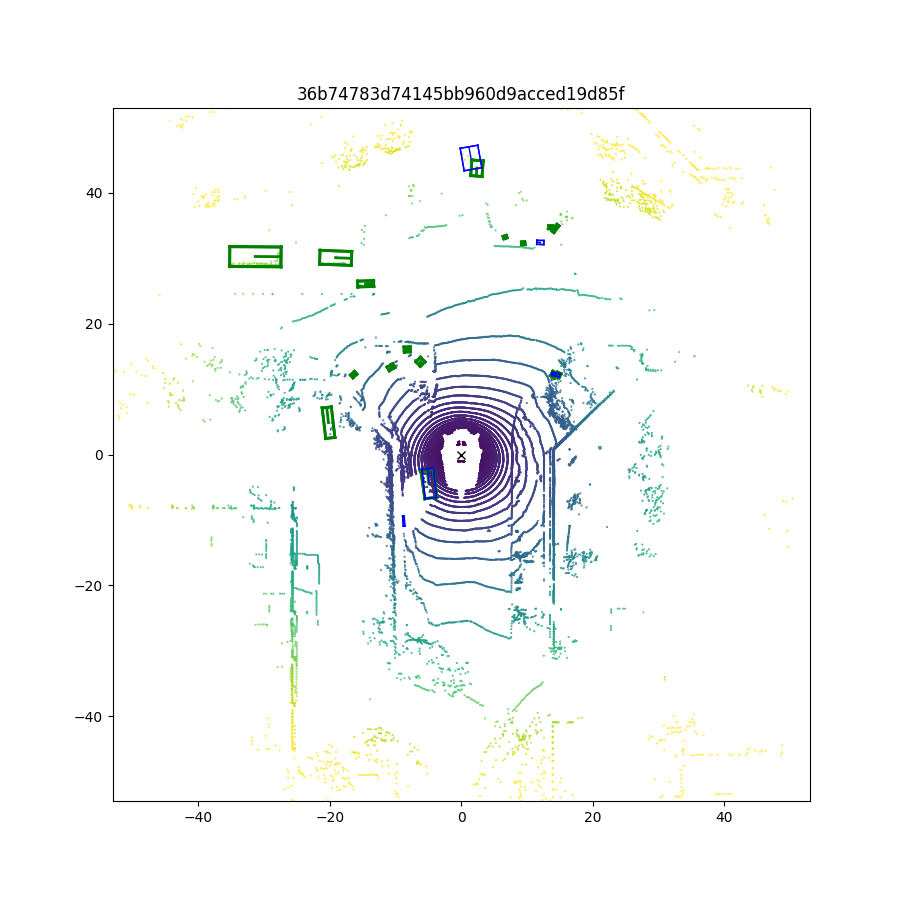}
    \includegraphics[width=0.48\linewidth]{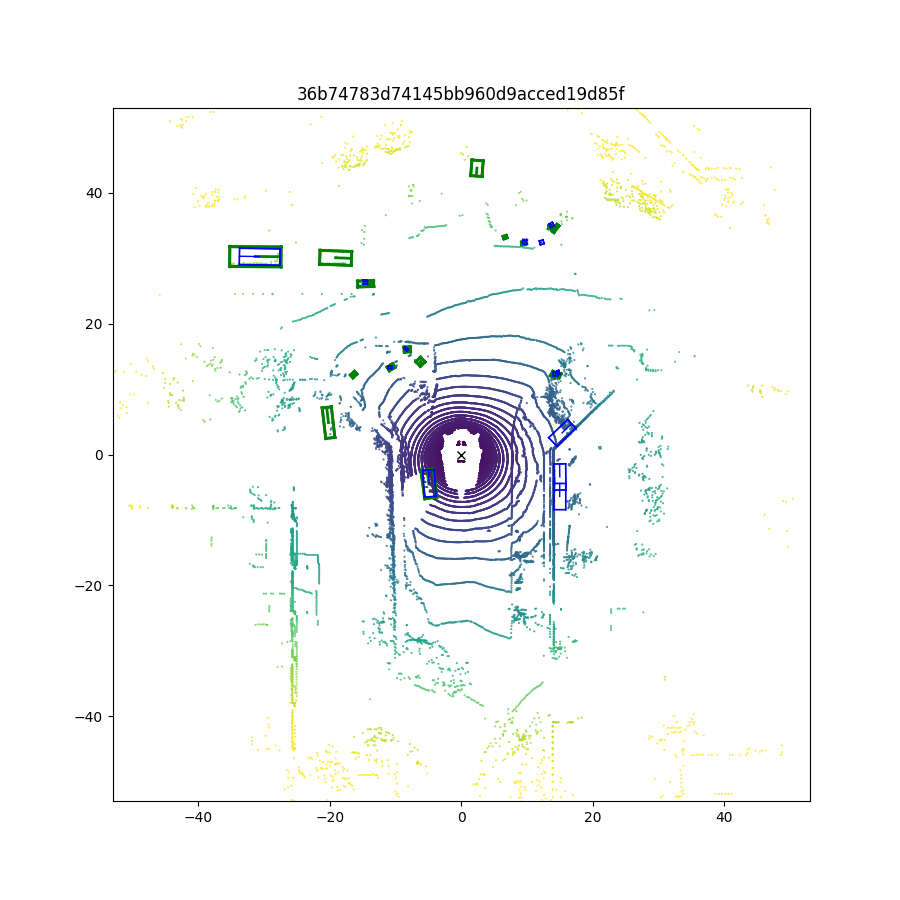}
    \caption[Qualitative comparison between UNION and  VESPA-Direct 4]{Qualitative comparison between UNION (left) and  \gls{vespa}-Direct (right). The scene in Figure~\ref{fig:qual_4} demonstrates  \gls{vespa}-Direct's strength in small object recall. It detects several small vehicles missed by UNION, though it also introduces more false positives. UNION shows higher precision, but its predictions are sparse and less complete.}
    \label{fig:qual_4}
\end{figure}

\begin{figure}[h]
    \centering
    \includegraphics[width=0.48\linewidth]{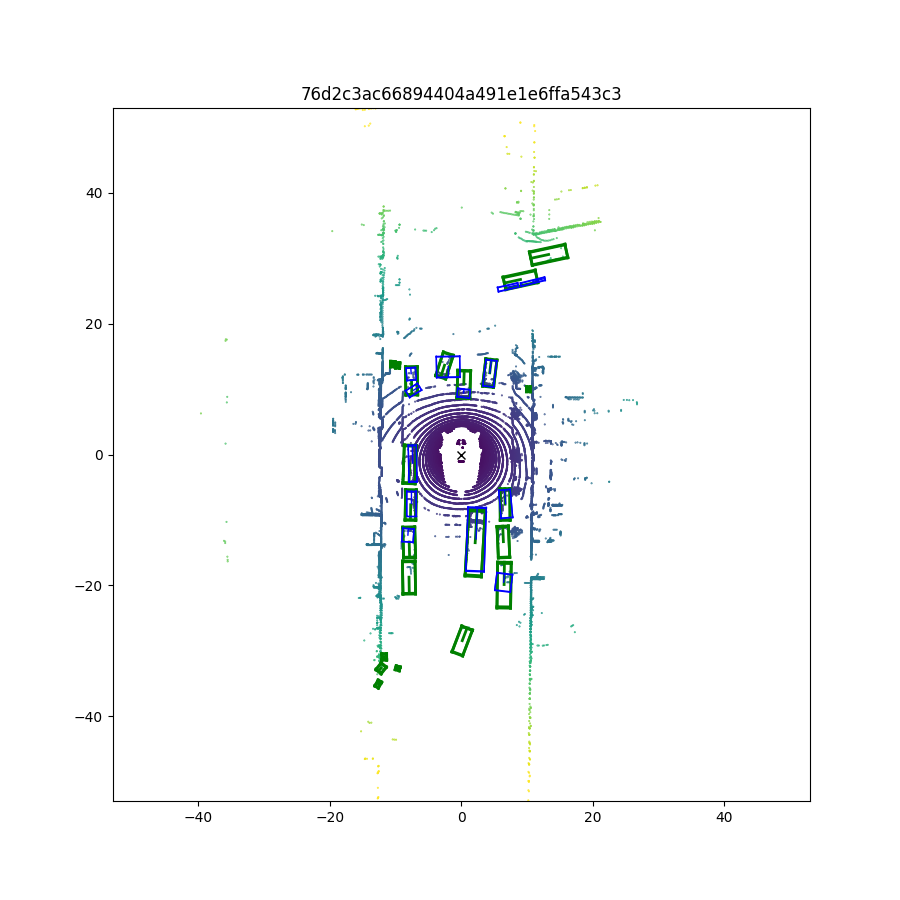}
    \includegraphics[width=0.48\linewidth]{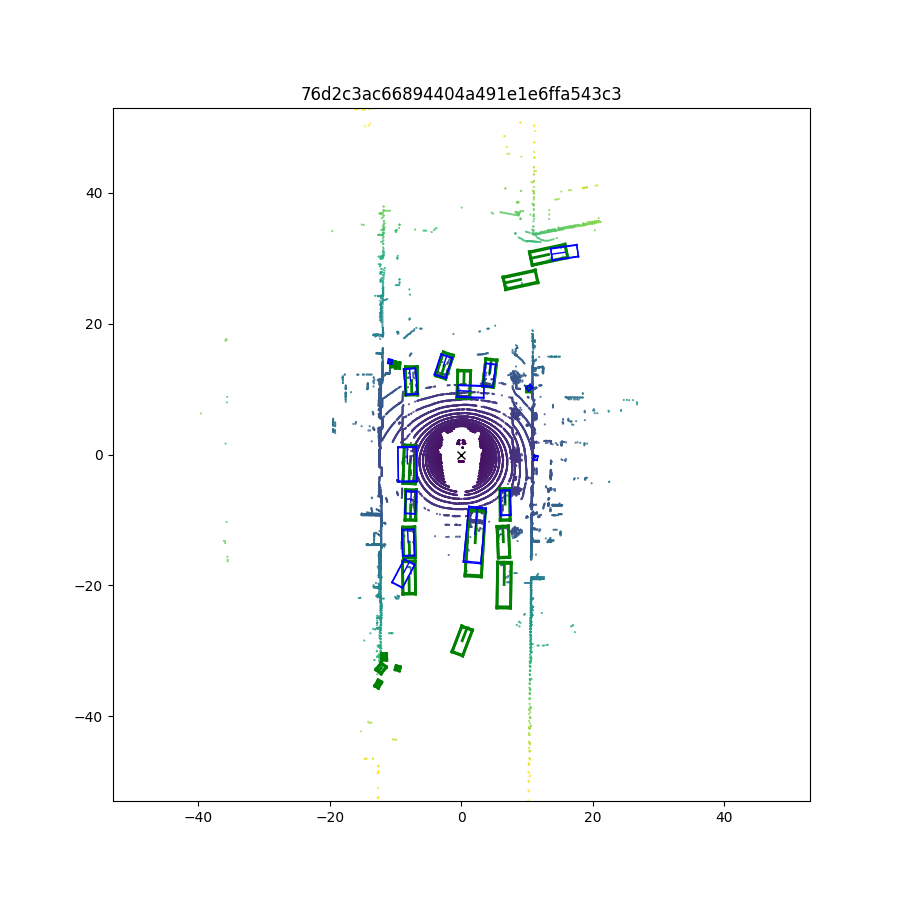}
    \caption[Qualitative comparison between UNION and  VESPA-Direct 6]{Qualitative comparison between UNION (left) and  \gls{vespa}-Direct (right). In Figure~\ref{fig:qual_6}, both methods perform similarly on car-like objects.  \gls{vespa}-Direct detects two additional small vehicles that UNION misses and offers slightly more accurate bounding boxes in some cases.}
    \label{fig:qual_6}
\end{figure}

\begin{figure}[h]
    \centering
    \includegraphics[width=0.48\linewidth]{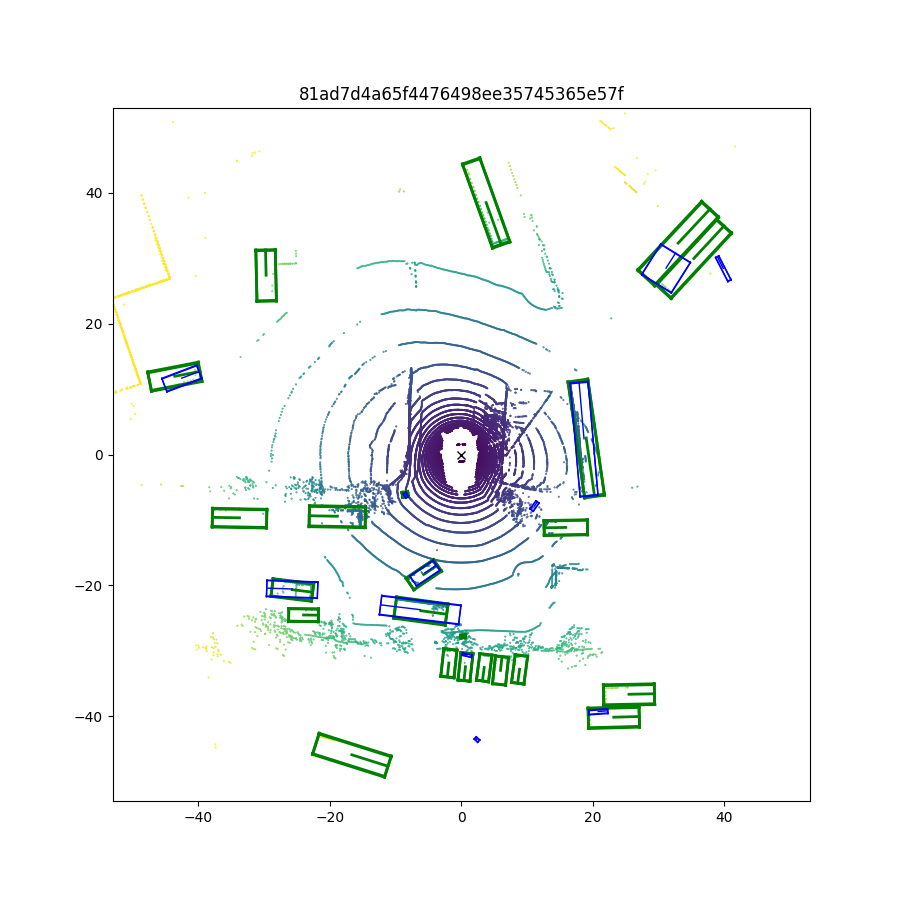}
    \includegraphics[width=0.48\linewidth]{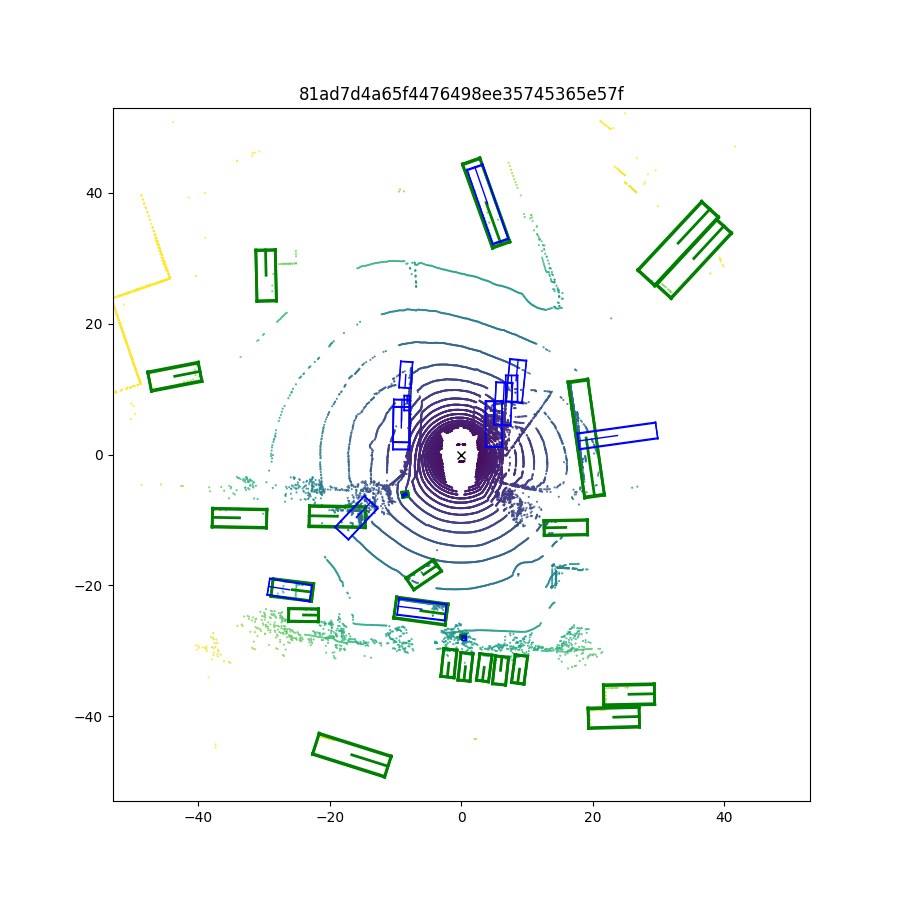}
    \caption[Qualitative comparison between UNION and  VESPA-Direct 7]{Qualitative comparison between UNION (left) and  \gls{vespa}-Direct (right). The example in Figure~\ref{fig:qual_7} shows  \gls{vespa}-Direct detecting three true positives with high accuracy. One orientation estimate is off by 90 degrees, and several false positives appear near a vehicle. UNION produces much fewer false positives in this case.}
    \label{fig:qual_7}
\end{figure}

\begin{figure}[h]
    \centering
    \includegraphics[width=0.48\linewidth]{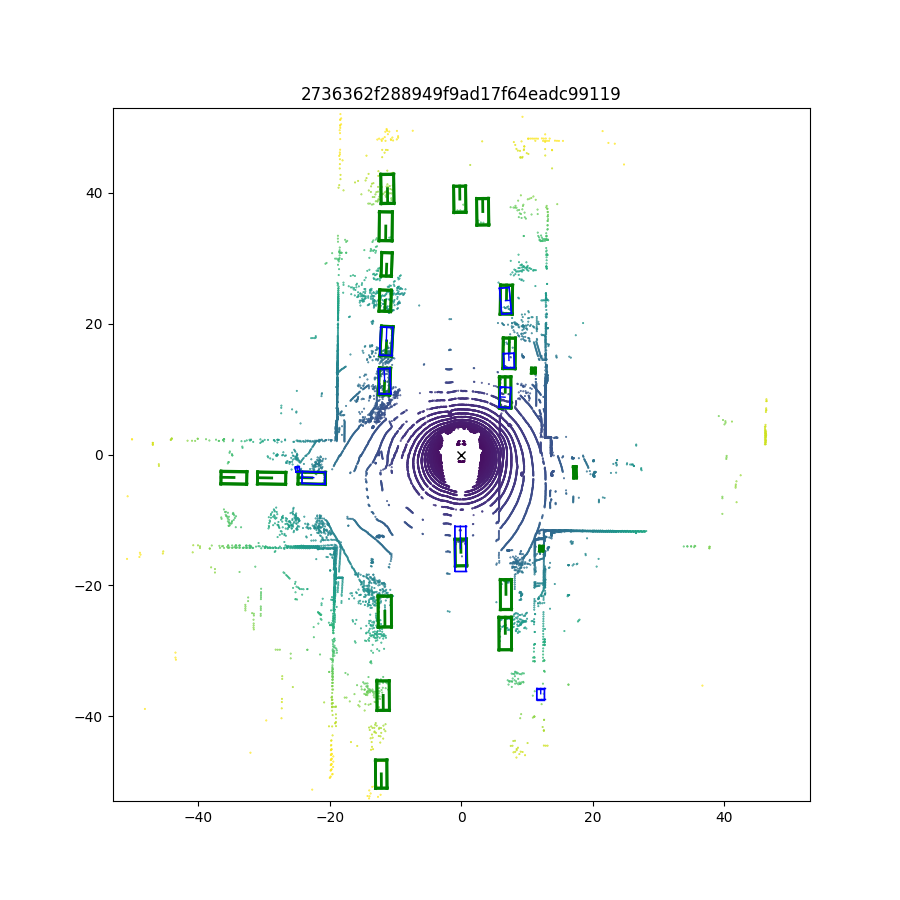}
    \includegraphics[width=0.48\linewidth]{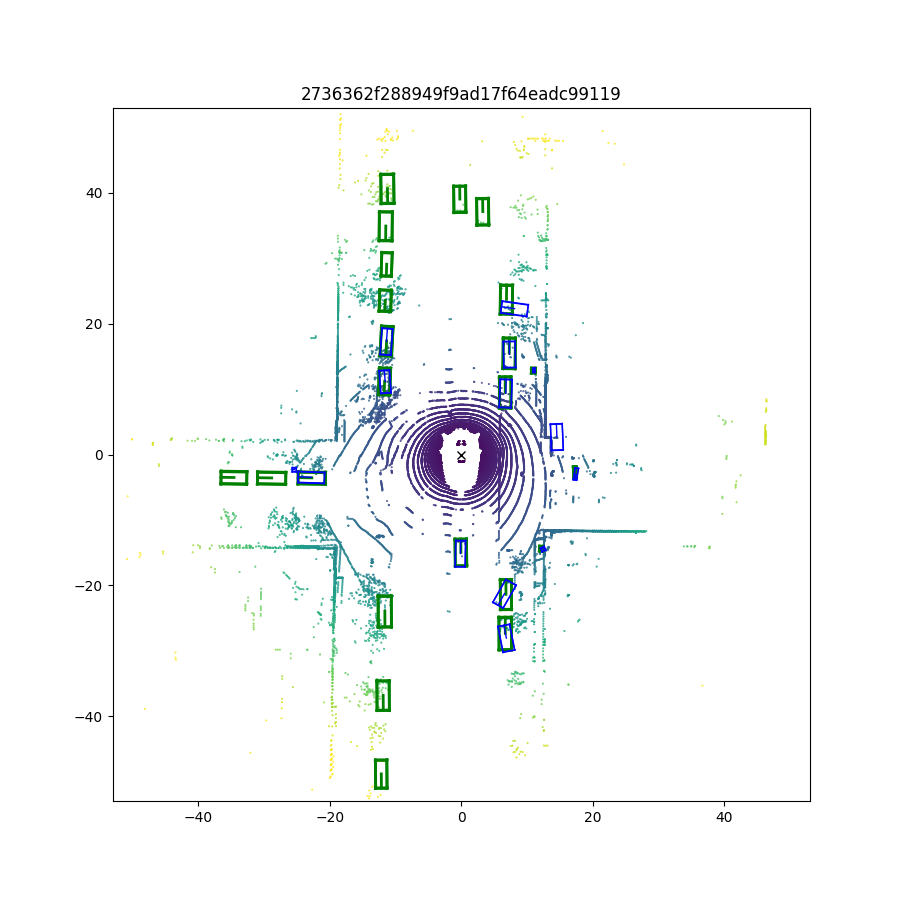}
    \caption[Qualitative comparison between UNION and  VESPA-Direct 8]{Qualitative comparison between UNION (left) and  \gls{vespa}-Direct (right). In Figure~\ref{fig:qual_8},  \gls{vespa}-Direct achieves significantly better recall, detecting all four small objects and two additional large ones. Most of its boxes, especially those closer to the ego vehicle, are well estimated. One box shows a 90-degree orientation error. UNION detects only one of the small objects and provides less accurate estimates overall.}
    \label{fig:qual_8}
\end{figure}

\begin{figure}[h]
    \centering
    \includegraphics[width=0.48\linewidth]{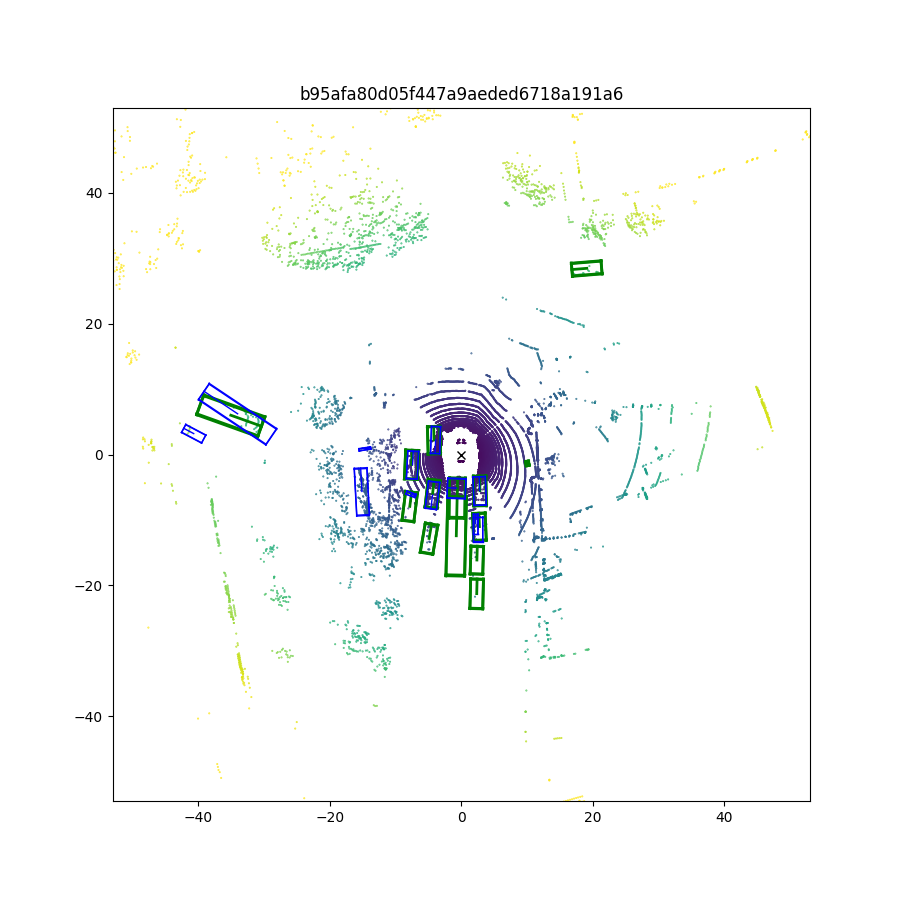}
    \includegraphics[width=0.48\linewidth]{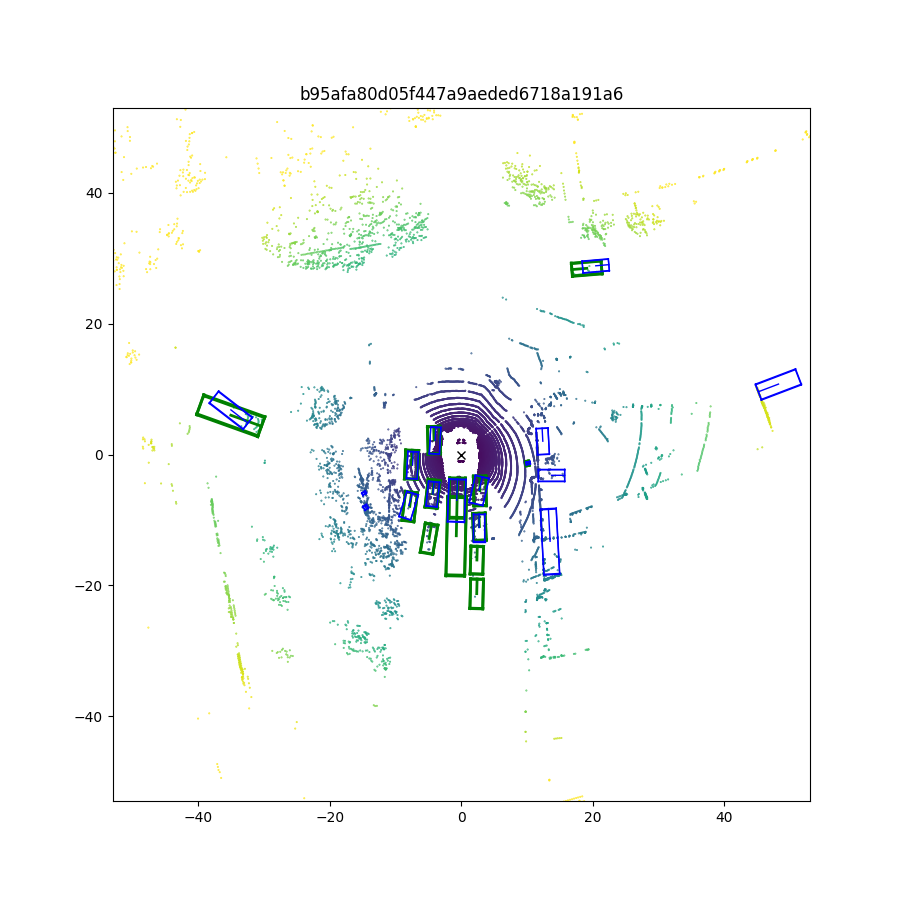}
    \caption[Qualitative comparison between UNION and  VESPA-Direct 9]{Qualitative comparison between UNION (left) and  \gls{vespa}-Direct (right). Figure~\ref{fig:qual_9} shows  \gls{vespa}-Direct successfully identifying a distant object missed by UNION, while also capturing a small nearby one. It produces several false positives, though its total true positive count slightly exceeds UNION’s.}
    \label{fig:qual_9}
\end{figure}

\begin{figure}[h]
    \centering
    \includegraphics[width=0.48\linewidth]{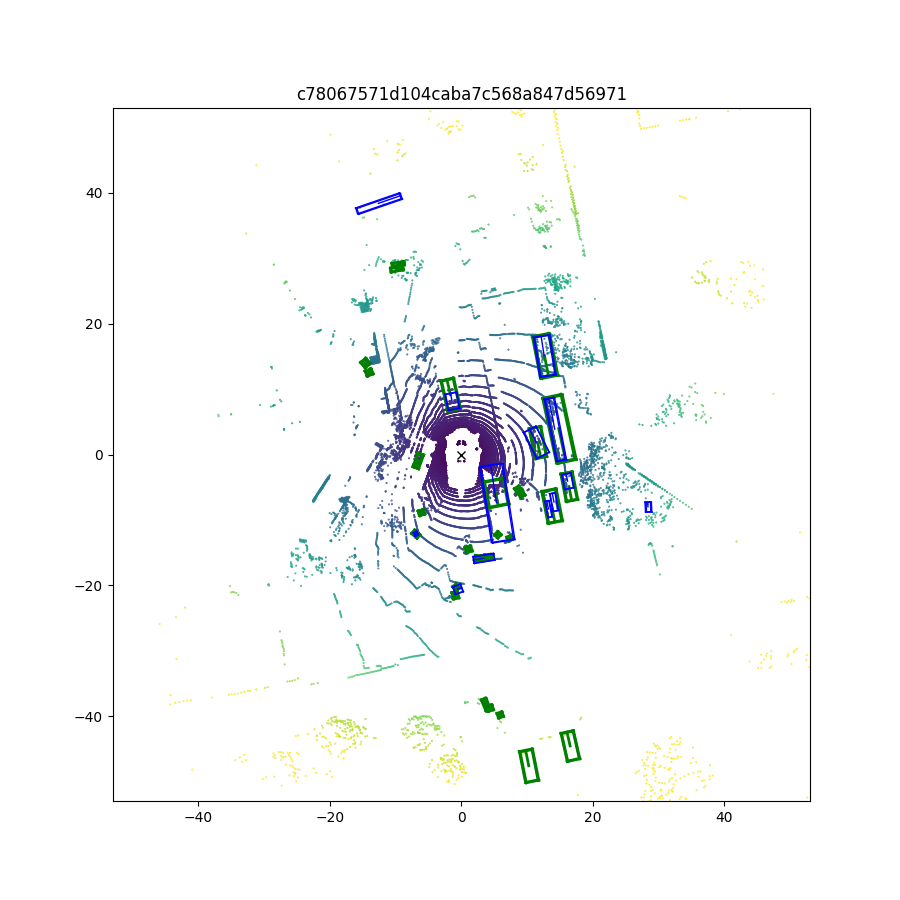}
    \includegraphics[width=0.48\linewidth]{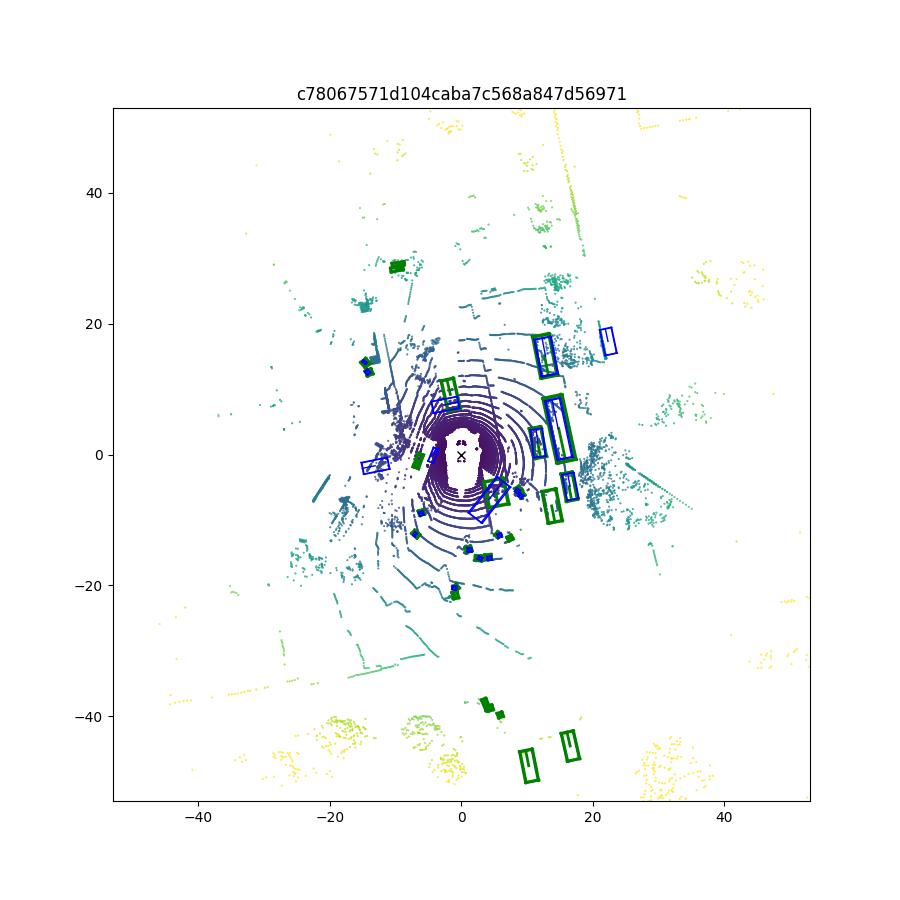}
    \caption[Qualitative comparison between UNION and  VESPA-Direct 10]{Qualitative comparison between UNION (left) and  \gls{vespa}-Direct (right). The next example (Figure~\ref{fig:qual_10}) shows  \gls{vespa}-Direct again outperforming UNION in small object detection. It identifies nearly all of them while UNION captures only a few. Most large object estimates by  \gls{vespa}-Direct are close to ground truth, though two are misoriented. UNION's estimates for large objects are less accurate regarding box sizes.}
    \label{fig:qual_10}
\end{figure}

\begin{figure}[h]
    \centering
    \includegraphics[width=0.48\linewidth]{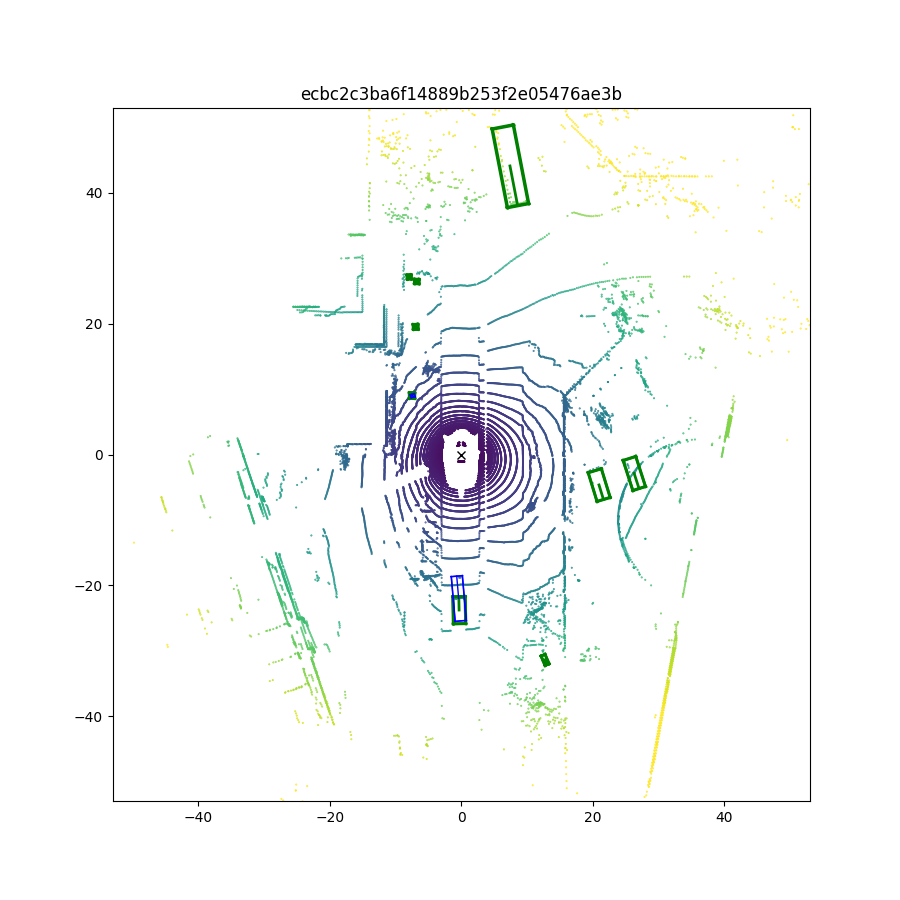}
    \includegraphics[width=0.48\linewidth]{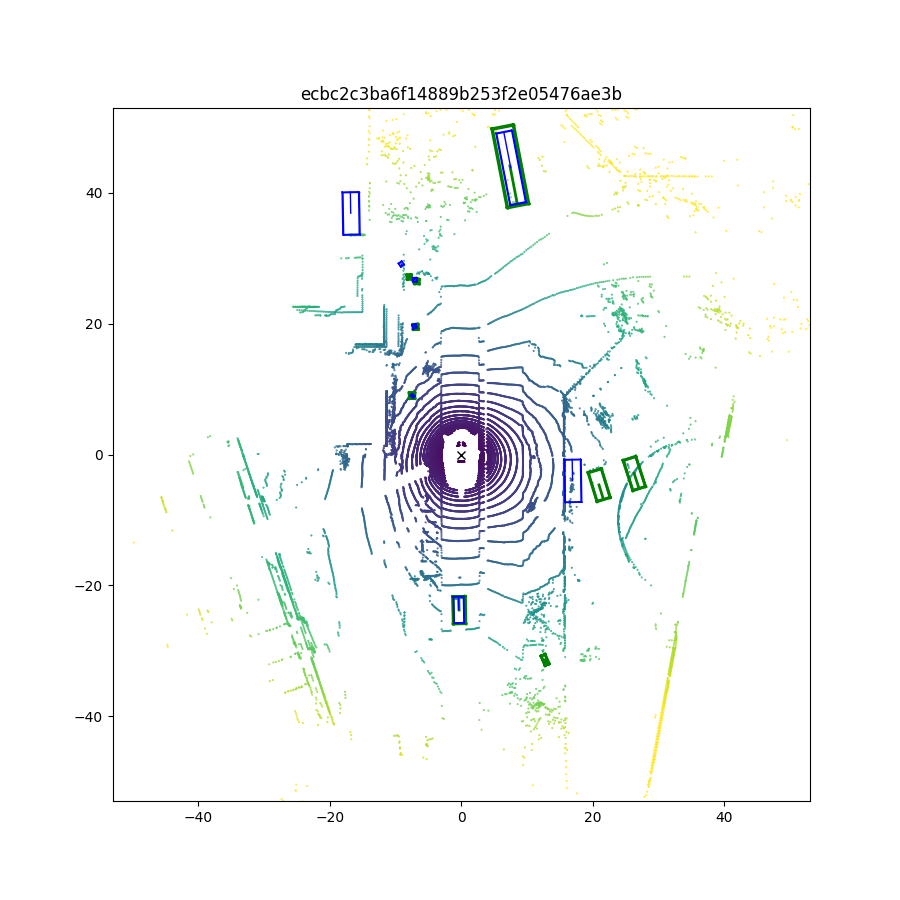}
    \caption[Qualitative comparison between UNION and  VESPA-Direct 11]{Qualitative comparison between UNION (left) and  \gls{vespa}-Direct (right). Finally, in Figure~\ref{fig:qual_11},  \gls{vespa}-Direct detects all four small objects, though one is slightly misplaced. It also captures a large distant vehicle with good box alignment, but also introduces 2 false positives. UNION misses three of the small objects and the distant vehicle. It detects a nearby car but overestimates its size.}
    \label{fig:qual_11}
\end{figure}

Across all examined scenes, \gls{vespa}-Direct consistently achieves higher recall, particularly in the detection of small and distant objects. Its bounding box estimates are often more precise in terms of size and orientation. However, it also introduces more false positives and occasional orientation errors. In contrast, UNION exhibits higher precision but often fails to capture smaller objects and struggles with box alignment. These results highlight  \gls{vespa}-Direct’s superior capability in dense or more complex environments with more difficult object classes.

\end{document}